\documentclass[11pt]{article}

\usepackage[preprint]{acl}

\usepackage{times}
\usepackage{latexsym}

\usepackage[T1]{fontenc}

\usepackage[utf8]{inputenc}

\usepackage{microtype}

\usepackage{inconsolata}

\usepackage{graphicx}

\usepackage{enumitem}

\usepackage{amsmath}
\usepackage{amssymb}

\usepackage{amsmath,amssymb}
\usepackage{bbm}
\usepackage{booktabs}
\usepackage[table]{xcolor}
\usepackage{colortbl}
\usepackage{multirow}
\usepackage{array}
\usepackage{pgf}
\usepackage{pifont}
\usepackage{graphicx}

\definecolor{goodcol}{HTML}{2D6A1E}
\definecolor{midcol}{HTML}{FDD835}
\definecolor{badcol}{HTML}{C62828}

\usepackage{tcolorbox}
\tcbuselibrary{skins,breakable}
\usepackage{graphicx}
\usepackage{pifont}
\usepackage{xcolor}
\usepackage{amssymb}

\definecolor{E3H}     {HTML}{7B2D8B}
\definecolor{E3Light} {HTML}{C89FD8}
\definecolor{E1H}     {HTML}{1558A0}
\definecolor{E1Light} {HTML}{92B8DC}
\definecolor{E2H}     {HTML}{A63D00}
\definecolor{E2Light} {HTML}{DFA882}

\definecolor{VidH}    {HTML}{1A6335}
\definecolor{AudH}    {HTML}{10456B}
\definecolor{TxtH}    {HTML}{7A3300}
\definecolor{VidBg}   {HTML}{F2FAF4}
\definecolor{AudBg}   {HTML}{EEF4FA}
\definecolor{TxtBg}   {HTML}{FDF5EE}

\definecolor{CleanG}  {HTML}{1A7A3A}
\definecolor{CorruptR}{HTML}{B81C1C}
\definecolor{GTGreen} {HTML}{145C2A}
\definecolor{PredRed} {HTML}{8B0000}
\definecolor{NeutralG}{HTML}{4A5568}

\definecolor{OutputBg}{HTML}{FFFDF0}
\definecolor{FailBg}  {HTML}{FFF0F0}
\definecolor{MetaBg}  {HTML}{F0F2F5}
\definecolor{OptionBg}{HTML}{FAFBFC}

\newcommand{\qcmark}{\textcolor{GTGreen}{\ding{51}}}
\newcommand{\qxmark}{\textcolor{PredRed}{\ding{55}}}

\newcommand{\cleanpill}{%
  \tcbox[on line, arc=2pt, boxrule=0pt, boxsep=0pt,
    left=3pt, right=3pt, top=1.5pt, bottom=1.5pt,
    colback=CleanG, colframe=CleanG]{%
    \color{white}\sffamily\bfseries\fontsize{6.5}{7}\selectfont\ CLEAN\ }}

\newcommand{\corruptpill}{%
  \tcbox[on line, arc=2pt, boxrule=0pt, boxsep=0pt,
    left=3pt, right=3pt, top=1.5pt, bottom=1.5pt,
    colback=CorruptR, colframe=CorruptR]{%
    \color{white}\sffamily\bfseries\fontsize{6.5}{7}\selectfont\ CORRUPT\ }}

\newcommand{\optGT}   [2]{\textbf{\textcolor{GTGreen}{#1.~#2~\qcmark}}}
\newcommand{\optPred} [2]{\textbf{\textcolor{PredRed}{#1.~#2~\qxmark}}}
\newcommand{\optOther}[2]{\textcolor{NeutralG}{#1.~#2}}
\newcommand{\optAbs}  [2]{\textit{\textcolor{NeutralG}{#1.~#2~{\fontsize{5.5}{6}\selectfont[abstain]}}}}

\newcommand{\metabadge}[1]{%
  \tcbox[on line, arc=2pt, boxrule=0.4pt, boxsep=0pt,
    left=3pt, right=3pt, top=1.5pt, bottom=1.5pt,
    colback=MetaBg, colframe=gray!35]{%
    \sffamily\fontsize{6.5}{7}\selectfont\textcolor{NeutralG}{#1}}}

\newcommand{\qualboxbegin}[4]{%
\begin{tcolorbox}[
  enhanced, breakable,
  arc=7pt, outer arc=7pt, boxrule=1.2pt,
  colframe=#1!60!black, colback=white,
  fuzzy shadow={1pt}{-1pt}{0pt}{0.3pt}{black!15},
  title={%
    \tcbox[on line, arc=3pt, boxrule=0.7pt, boxsep=0pt,
      left=5pt, right=5pt, top=2.5pt, bottom=2.5pt,
      colback=#2!60!white, colframe=white!50!#1]{%
      \sffamily\bfseries\fontsize{8.5}{10}\selectfont\color{#1!20!black}\ #3\ }%
    \enspace
    {\sffamily\bfseries\fontsize{8.5}{10}\selectfont\color{white}#4}},
  attach boxed title to top left={yshift=-2.8mm, xshift=9pt},
  boxed title style={
    enhanced,
    interior style={
      left color=#1!30!white,
      right color=#1!90!black,
    },
    colframe=#1!65!black,
    arc=5pt, outer arc=5pt,
    boxrule=0.8pt,
    fuzzy shadow={0.5pt}{-0.5pt}{0pt}{0.2pt}{black!20},
  },
  top=5pt, bottom=7pt, left=7pt, right=7pt,
  before upper={\setlength{\parskip}{0pt}\setlength{\parindent}{0pt}},
]}
\newcommand{\qualboxend}{\end{tcolorbox}}

\tcbset{
  vbox/.style={
    enhanced, arc=5pt, outer arc=5pt, boxrule=0.6pt,
    colframe=VidH!70!black, colback=VidBg,
    interior style={top color=VidBg, bottom color=VidBg!95!VidH},
    colbacktitle=VidH, coltitle=white,
    fonttitle=\sffamily\bfseries\fontsize{7.5}{8.5}\selectfont,
    title={$\blacksquare$\ \ VIDEO},
    toptitle=2.5pt, bottomtitle=2.5pt,
    top=5pt, bottom=4pt, left=4pt, right=4pt,
    before upper={\setlength{\parskip}{2pt}\setlength{\parindent}{0pt}},
  },
  abox/.style={
    enhanced, arc=5pt, outer arc=5pt, boxrule=0.6pt,
    colframe=AudH!70!black, colback=AudBg,
    interior style={top color=AudBg, bottom color=AudBg!95!AudH},
    colbacktitle=AudH, coltitle=white,
    fonttitle=\sffamily\bfseries\fontsize{7.5}{8.5}\selectfont,
    title={$\blacktriangleright$\ \ AUDIO},
    toptitle=2.5pt, bottomtitle=2.5pt,
    top=5pt, bottom=4pt, left=4pt, right=4pt,
    before upper={\setlength{\parskip}{2pt}\setlength{\parindent}{0pt}},
  },
  tbox/.style={
    enhanced, arc=5pt, outer arc=5pt, boxrule=0.6pt,
    colframe=TxtH!70!black, colback=TxtBg,
    interior style={top color=TxtBg, bottom color=TxtBg!95!TxtH},
    colbacktitle=TxtH, coltitle=white,
    fonttitle=\sffamily\bfseries\fontsize{7.5}{8.5}\selectfont,
    title={$\equiv$\ \ TEXT},
    toptitle=2.5pt, bottomtitle=2.5pt,
    top=5pt, bottom=4pt, left=4pt, right=4pt,
    before upper={\setlength{\parskip}{2pt}\setlength{\parindent}{0pt}},
  },
  obox/.style={
    enhanced, arc=5pt, boxrule=0.5pt,
    colframe=gray!25, colback=OutputBg,
    top=4pt, bottom=4pt, left=5pt, right=5pt,
    before upper={\setlength{\parskip}{2pt}\setlength{\parindent}{0pt}},
  },
  fbox2/.style={
    enhanced, arc=5pt, boxrule=0.5pt,
    colframe=PredRed!20, colback=FailBg,
    top=4pt, bottom=4pt, left=5pt, right=5pt,
    before upper={\setlength{\parskip}{2pt}\setlength{\parindent}{0pt}},
  },
  gridwrap/.style={
    enhanced, sharp corners, boxrule=0pt,
    colback=white, colframe=white,
    left=0pt, right=0pt, top=0pt, bottom=0pt,
    sidebyside, sidebyside align=top seam,
    lefthand ratio=0.33, sidebyside gap=6pt,
  },
}

\newcommand{\modalitygrid}[7]{%
\begin{tcolorbox}[gridwrap]
  \begin{tcolorbox}[vbox, width=\linewidth]
    #1\par\vspace{4pt}%
    \begin{tcolorbox}[enhanced, arc=4pt, boxrule=0.5pt,
        colframe=VidH!30!gray, colback=gray!4,
        top=0pt, bottom=0pt, left=0pt, right=0pt,
        before upper={}]
      \centering #2%
    \end{tcolorbox}%
    \vspace{3pt}{\fontsize{7}{8.5}\selectfont\textit{#7}}%
  \end{tcolorbox}%
  \tcblower
  \begin{tcolorbox}[abox, width=\linewidth]
    #3\par\vspace{3pt}%
    {\fontsize{8}{10}\selectfont #4}%
  \end{tcolorbox}%
  \vspace{5pt}%
  \begin{tcolorbox}[tbox, width=\linewidth]
    #5\par\vspace{3pt}%
    {\fontsize{8}{10}\selectfont #6}%
  \end{tcolorbox}%
\end{tcolorbox}}

\newcommand{\optionblock}[1]{%
\begin{tcolorbox}[enhanced, arc=5pt, boxrule=0.5pt,
    colframe=gray!20, colback=OptionBg,
    top=5pt, bottom=5pt, left=6pt, right=6pt,
    before upper={\setlength{\parskip}{0pt}\setlength{\parindent}{0pt}}]
#1
\end{tcolorbox}}

 \usepackage{booktabs}
 \usepackage{colortbl}
 \usepackage{xcolor}
 \usepackage{multirow}
 \usepackage{makecell}
 \usepackage{graphicx}  
 \usepackage{array}
 \usepackage{siunitx}   

\definecolor{abs0}{HTML}{F7FCF5}      
\definecolor{abs1}{HTML}{C7E9C0}      
\definecolor{abs2}{HTML}{74C476}      
\definecolor{abs3}{HTML}{41AB5D}      
\definecolor{abs98}{HTML}{CB181D}     
\definecolor{abs97}{HTML}{EF3B2C}     
\definecolor{rowshade}{HTML}{F5F5F5}  
\definecolor{headgray}{HTML}{4A4A4A}  

\newcommand{\abscell}[2]{\cellcolor{#1}\textbf{#2}}

\newcommand{\gc}[2]{%
  \pgfmathtruncatemacro{\mval}{min(max(#2, 0), 100)}%
  \ifnum\mval>50
    \pgfmathtruncatemacro{\mixpct}{min(round((\mval - 50) * 2), 100)}%
    \edef\doacc{\noexpand\cellcolor{goodcol!\mixpct!midcol}{#1}}%
  \else
    \pgfmathtruncatemacro{\mixpct}{min(round(\mval * 2), 100)}%
    \edef\doacc{\noexpand\cellcolor{midcol!\mixpct!badcol}{#1}}%
  \fi
  \doacc
}

\newcommand{\cmark}{\ding{51}}
\newcommand{\xmark}{\ding{55}}

\newcommand{\sv}[2]{#1\,{\tiny\textcolor{gray}{#2\%}}}

\title{Omni-Modal Dissonance Benchmark: Systematically Breaking Modality Consensus to Probe Robustness and Calibrated Abstention\vspace{5pt}}

\author{
\vspace{10pt}
    Zabir Al Nazi\hspace{.1em}$^{\,1,\ast}$,\quad
    Shubhashis Roy Dipta\hspace{.1em}$^{\,2,\ast}$,\quad
    Md Rizwan Parvez\hspace{.1em}$^{\,3}$ \vspace{1mm}\\ 
    \vspace{2pt}
    $^1$\textbf{University of California, Riverside} \\
    $^2$\textbf{University of Maryland, Baltimore County}\\
    $^3$\textbf{Qatar Computing Research Institute (QCRI)}\vspace{1mm}\\
    \texttt{znazi002@ucr.edu}, \texttt{sroydip1@umbc.edu}, \texttt{mparvez@hbku.edu.qa}\\
    {\scriptsize $^\ast$Equal contribution}
}

\begin{document}
\maketitle

\begin{abstract}
  Existing omni-modal benchmarks attempt to measure modality-specific contributions, but their measurements are \textit{confounded}: naturally co-occurring modalities carry correlated yet unequal information, making it unclear whether results reflect true modality reliance or information asymmetry. We introduce OMD-Bench, where all modalities are initially congruent - each presenting the same anchor, an object or event independently perceivable through video, audio, and text - which we then systematically corrupt to isolate each modality's contribution. We also evaluate calibrated abstention: whether models appropriately refrain from answering when evidence is conflicting. The benchmark comprises 4,080 instances spanning 27 anchors across eight corruption conditions. Evaluating ten omni-modal models under zero-shot and chain-of-thought prompting, we find that models over-abstain when two modalities are corrupted yet under-abstain severely when all three are, while maintaining high confidence (${\sim}$60--100\%) even under full corruption. Chain-of-thought prompting improves abstention alignment with human judgment but amplifies overconfidence rather than mitigating it.  OMD-Bench provides a diagnostic benchmark for diagnosing modality reliance, robustness to cross-modal inconsistency, and uncertainty calibration in omni-modal systems.
\end{abstract}
\section{Introduction}
\label{sec:intro}


Omni-modal models - systems that jointly process video, audio, and text within a single architecture - have become central to multimodal AI \citep{openai2024gpt4o, team2024gemini, xu2025qwen25omni, fu2024vita}. By encoding all three signals together, these models aim for richer understanding and more robust reasoning than bimodal predecessors \citep{liu2024llava, dai2023instructblip, chu2023qwenaudio}. Yet a fundamental diagnostic question remains open: \textbf{\textit{how do these models actually use each modality, and what happens when modalities disagree?}}

\begin{minipage}[!t]{0.48\textwidth}
  \centering
  \includegraphics[width=\linewidth]{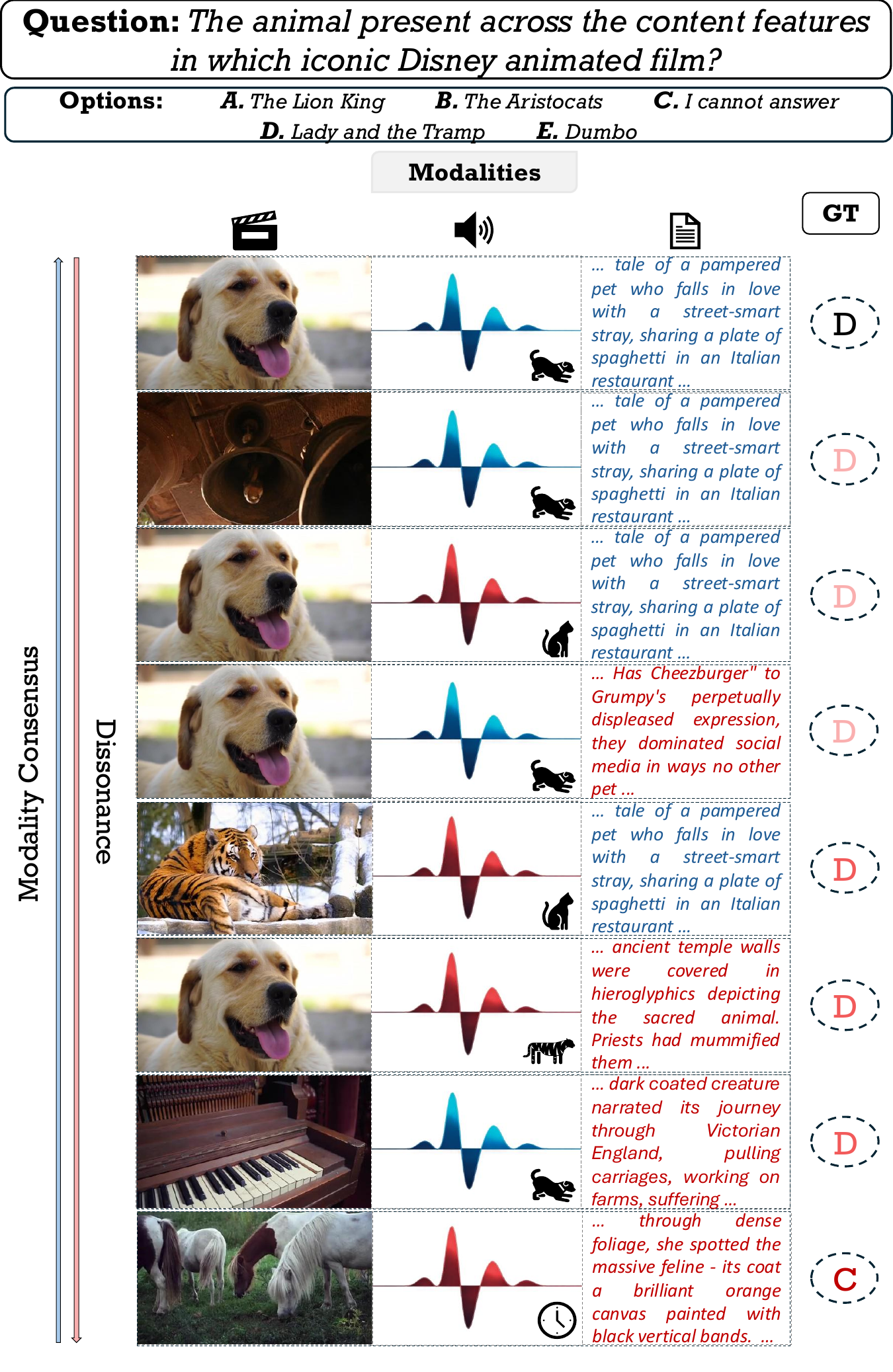}
  \captionof{figure}{\textbf{Overview of OMD-Bench.} All modalities initially depict the same anchor (a dog). We systematically corrupt subsets of modalities, replacing them with different anchor's content, and evaluate whether models can still answer correctly or appropriately abstain.}
  \label{fig:intro}
\end{minipage}

This question matters in real-world, where inputs are often noisy, incomplete, or contradictory - a surveillance system may receive garbled audio alongside clear video, or a medical assistant may encounter clinical notes that conflict with imaging data. A model that silently over-relies on one modality, or that answers confidently under contradictory evidence, poses serious reliability risks.

Several benchmarks evaluate omni-modal models along these lines. OmniBench \citep{li2024omnibench}, AV-Odyssey \citep{gong2024avodyssey}, WorldSense \citep{hong2025worldsense}, and UNO-Bench \citep{chen2025unobench} test joint reasoning over image or video, audio, and text, while work on inconsistency reasoning \citep{yan2025mmir} and modality bias \citep{chen2024quantifying, zheng2025modality} shows that models often fail under cross-modal conflict. However, these evaluations share a fundamental \textbf{confound}: naturally co-occurring modalities carry \textit{correlated but unequal} information, making it hard to determine whether results reflect genuine cross-modal integration or reliance on a dominant modality. Work on missing modalities \citep{baltrusaitis2019multimodal} addresses robustness to absent inputs, but absence differs from contradiction - a model may handle a missing modality gracefully yet fail when it actively conveys conflicting information.

We argue that what is needed is a \textit{controlled perturbation} 
approach: rather than passively observing performance on naturally 
occurring data, we systematically replace subsets of modalities 
while tracking the effect on model predictions. Drawing on the 
logic of interventional reasoning~\citep{pearl2009causality}, 
this requires a property that existing benchmarks lack: all 
modalities must initially encode the \textit{same} underlying concept 
with comparable informativeness, so that any perturbation produces 
an interpretable signal.

We introduce \textbf{OMD-Bench} (Omni-Modal Dissonance Benchmark), organized around 27 \textit{anchors} - objects or events (e.g., a dog barking, a bell ringing) that are independently perceivable through video, audio, and text. Each sample begins fully congruent, with all three modalities presenting the same anchor. We then apply a \textbf{factorial corruption protocol}: for each of the 255 base question - answer pairs, we generate all $2^3 = 8$ conditions by replacing subsets of modalities with content from a different anchor, totaling 4,080 instances across real and synthetic splits.

This design supports two analyses that prior benchmarks do not. \textbf{First}, because the clean condition provides a per-sample baseline and corruption targets one modality at a time, we can attribute each modality's contribution with substantially less confounding. \textbf{Second}, human annotators determined per-instance whether the available modalities suffice to answer or whether abstention is appropriate, enabling evaluation of calibrated abstention - a capability studied in text-only LLMs \citep{wen2024abstention} but largely unexplored in the omni-modal setting.


We evaluate ten state-of-the-art omni-modal models under both zero-shot and chain-of-thought prompting. Our main findings are:

\begin{itemize}[leftmargin=*, itemsep=2pt, topsep=2pt]
  \item \textbf{Asymmetric modality reliance.} Most models disproportionately rely on text even when video and audio carry equally informative content, consistent with findings from bimodal settings \citep{chen2024quantifying, zheng2025modality} and now quantified through controlled intervention in the tri-modal case.

  \item \textbf{Poor abstention calibration.} Models over-abstain when two modalities are corrupted yet the remaining modality provides sufficient evidence, and severely under-abstain when all three are corrupted and abstention is warranted, revealing miscalibration in both directions.

  \item \textbf{Fragility to cross-modal conflict.} Accuracy degrades sharply under corruption, with disproportionate drops depending on which modality is corrupted, indicating limited robustness to cross-modal inconsistencies.

  \item \textbf{Chain-of-thought presents a calibration trade-off.} 
CoT prompting improves both clean accuracy and abstention alignment 
with human judgment for most models, but amplifies overconfidence 
even on incorrect predictions.
\end{itemize}

In summary, our contributions are:
\begin{enumerate}[leftmargin=*, itemsep=2pt, topsep=2pt]
  \item \textbf{OMD-Bench}, a benchmark that uses controlled factorial intervention on initially congruent tri-modal inputs to enable less confounded attribution of modality reliance, addressing a limitation shared by prior omni-modal benchmarks.

  \item A formalization and evaluation of \textbf{calibrated abstention} in the omni-modal setting, providing systematic measurement of whether models appropriately refrain from answering under cross-modal conflict.

  \item \textbf{Knowledge-grounded questions} with balanced distractors that require both cross-modal perception and world-knowledge retrieval.

  \item An \textbf{empirical analysis of ten omni-modal models}, revealing systematic patterns of modality bias, abstention failure, and overconfidence with implications for reliable deployment.
\end{enumerate}

\section{Related Work}
\label{sec:related}

\subsection{Omni-Modal Benchmarks}

Several benchmarks evaluate models across three or more modalities: OmniBench \citep{li2024omnibench} tests joint reasoning over image, audio, and text; AV-Odyssey \citep{gong2024avodyssey} targets audio-visual perception; WorldSense \citep{hong2025worldsense} probes temporally grounded omni-modal understanding; and UNO-Bench \citep{chen2025unobench} studies the relationship between unimodal and omni-modal performance.
These benchmarks rely on naturally co-occurring data where modalities carry correlated but unequal information, making it difficult to distinguish genuine cross-modal integration from reliance on a dominant modality. OMD-Bench addresses this by constructing initially congruent tri-modal inputs and applying factorial corruption.

\subsection{Modality Bias \& Cross-Modal Robustness}

\citet{chen2024quantifying} show that multimodal LLMs exhibit strong unimodal biases, often defaulting to language shortcuts. \citet{zheng2025modality} confirm that this bias persists in recent systems. \citet{yan2025mmir} reveal that models struggle when text and image conflict, though their benchmark (MMIR) is limited to the bimodal setting.
Work on missing modalities \citep{baltrusaitis2019multimodal, wu2024missing} addresses robustness to absent inputs. OMD-Bench targets a complementary problem: robustness to \textit{contradictory} modalities, where a channel actively conveys conflicting rather than missing information.

\subsection{Selective Prediction \& Abstention}

Selective prediction and abstention are well-studied in text-only settings \citep{wen2024abstention, geifman2017selective}. \citet{chen2025unveiling} extend calibration analysis to multimodal LLMs, finding systematic miscalibration and a tendency to answer rather than abstain. However, calibrated abstention under controlled cross-modal conflict in the omni-modal setting remains unexplored. OMD-Bench fills this gap with an explicit abstention option and human-annotated ground truth identifying when abstention is appropriate.


\section{OMD-Bench: Design and Construction}
\label{sec:benchmark}

We use \textit{omni-modal} to denote the joint processing of visual, auditory, and textual input - the three perceptual channels that current omni-modal MLLMs are designed to handle simultaneously~\citep{team2024gemini}.
OMD-Bench evaluates all three channels jointly, unlike vision-language or audio-language benchmarks that cover only a subset.

\subsection{Design Principles}
\label{sec:principles}

OMD-Bench is built on three principles:

\paragraph{Interventional control.}
Each modality can be independently replaced while holding the others constant~\citep{pearl2009causality}.
This requires that all modalities initially encode the same concept with approximately equal informativeness (\S\ref{sec:anchors}).

\paragraph{Factorial completeness.}
For each sample, we generate all $2^3{=}8$ corruption patterns, providing a complete picture of how each modality and every combination of modalities contributes to model predictions.

\paragraph{Evidence-grounded abstention.}
Every question includes an explicit abstention option.
The correct response for each instance is determined by human annotators following a structured evidence protocol (\S\ref{sec:eval-protocol}, Appendix~\ref{app:annotation}): when the available modalities provide sufficient and consistent evidence, the factual answer is correct; when the evidence is insufficient or contradictory, abstention is the appropriate response.
This per-instance grounding, motivated by selective prediction norms~\citep{geifman2017selective} and findings that MLLMs are severely overconfident under conflicting inputs~\citep{chen2025unveiling,yan2025multimodal}, enables principled measurement of whether models know when not to answer.

\subsection{Anchors: Tri-Modal Knowledge Units}
\label{sec:anchors}

The building block of OMD-Bench is the \textit{anchor}: a real-world entity or event that is independently perceivable through video, audio, and text.
We curate 27 anchors spanning animals (e.g., \textit{dog barking}), natural phenomena (e.g., \textit{thunder}), human activities (e.g., \textit{playing guitar}), and mechanical events (e.g., \textit{clock ticking}).

Each anchor must satisfy three selection criteria.
\textbf{(i)~Tri-modal recognizability}: the entity can be identified from any single modality alone, ensuring each channel is independently informative - a requirement analogous to the ``visual grounding'' constraint in VQA benchmarks~\citep{goyal2017making} but extended to all three modalities.
To verify equal informativeness empirically, human annotators identified anchors from each modality in isolation: recognition rates were 94.5\% (video), 94.4\% (text), and 93.1\% (audio) on the real split, and 95.2\%, 94.4\%, and 93.5\% respectively on the synthetic split, with a maximum cross-modality gap of 1.4\% (real) and 1.7\% (synthetic).
This confirms that no single modality is inherently easier to 
identify from than another in our dataset. Since answering requires first 
recognizing the anchor and then retrieving world knowledge 
(\S\ref{sec:anchors}), and the knowledge step is the same 
regardless of modality, equal recognition rates remove the 
informativeness confound present in prior benchmarks.
\textbf{(ii)~World-knowledge grounding}: the entity has associated world-knowledge facts (e.g., a guitar has six strings in standard tuning) that support non-trivial question-answer pairs, following the knowledge-grounded VQA paradigm~\citep{marino2019okvqa}.
\textbf{(iii)~Cross-anchor distinctiveness}: anchors are sufficiently dissimilar that replacing one anchor's modality content with another's creates unambiguous semantic conflict (\S\ref{sec:corruption}).

For each anchor, we construct knowledge-grounded questions requiring two cognitive steps: \textit{perceiving} which entity the modalities depict, then \textit{retrieving} world knowledge to answer.
Answering correctly requires both recognizing the anchor and knowing the relevant fact.
All questions and answer options were authored and validated by human annotators (details in Appendix~\ref{app:annotation}).

\subsection{Corruption Protocol}
\label{sec:corruption}

The corruption protocol enables controlled perturbation by 
systematically replacing modality channels.
Let $\mathbf{x} = (v, a, t)$ denote a tri-modal input depicting 
anchor~$\alpha$.
A corruption replaces one or more channels with content from 
different anchors~$\alpha' \neq \alpha$, where each corrupted 
modality's replacement anchor is sampled independently and 
uniformly from the remaining anchors.
The corruption vector $\mathbf{c} = (c_v, c_a, c_t) \in \{0,1\}^3$, 
where $c_m{=}0$ is original content and $c_m{=}1$ is replacement, 
specifies one of eight conditions ($C_{000}$ through $C_{111}$; 
see Table~\ref{tab:conditions} in Appendix~\ref{app:conditions}).
Since all three modalities initially depict the same 
anchor~$\alpha$, differences across conditions reflect only 
the corruption pattern.

This design supports two levels of analysis:
\textbf{Single-modality corruption} ($k{=}1$: conditions $C_{100}$, $C_{010}$, $C_{001}$) isolates each modality's contribution, with the accuracy drop from $C_{000}$ to $C_{m}$ quantifying reliance on modality~$m$.
\textbf{Majority corruption} ($k{\geq}2$: conditions $C_{110}$, $C_{101}$, $C_{011}$, $C_{111}$) creates inputs where most modalities point to the wrong or different anchors; abstention is the appropriate response for nearly all $k{=}3$ instances (${\approx}98\%$
human-annotated rate), while at $k{=}2$ the single remaining clean modality typically
suffices and the factual answer remains correct (${\leq}4\%$ abstention rate;
Table~\ref{tab:abstention-distribution}).

\subsection{Answer Option Design}
\label{sec:distractors}

Each question has five options: the correct answer for anchor~$\alpha$, three distractors that are correct answers to the \textit{same question} applied to different anchors, and an explicit abstention option.
This ensures all non-abstention options are plausible, so the question cannot be answered without first recognizing the anchor from the multimodal input~\citep{alhazmi2024distractor}.
Option order is randomized per instance to mitigate positional bias~\citep{pezeshkpour2024large}.

\subsection{Data Sourcing and Splits}
\label{sec:splits}

We construct two parallel splits to test generalization across data distributions following \cite{chen2024omnixr}. The \textbf{real} split uses videos from \href{https://www.pexels.com}{Pexels}, audio from \href{https://freesound.org}{Freesound} and \href{https://pixabay.com/sound-effects/}{Pixabay}, and human-written text descriptions. The \textbf{synthetic} split uses videos from generative models (\href{https://openai.com/sora}{Sora}, \href{https://x.ai/grok}{Grok}, \href{https://deepmind.google/models/gemini-image/}{Gemini}), audio from \href{https://github.com/zeyuet/AudioX}{AudioX}~\citep{tian2025audiox}, and same text descriptions. Text descriptions identify the anchor without revealing the answer to any associated question. All data underwent multi-stage human verification; details are in Appendix~\ref{app:annotation}.

The final benchmark comprises 27 anchors $\times$ ${\sim}$9.4 questions per anchor $=$ 255 seed questions, each evaluated under 8 corruption conditions, yielding \textbf{2,040} instances per split and \textbf{4,080} total.

\section{Experimental Setup}
\label{sec:setup}

\subsection{Models}

We evaluate ten omni-modal models capable of processing video, audio, and text simultaneously, spanning a range of scales, architectures, and access types to ensure our findings generalize beyond any single model family.
\paragraph{Open-weight models.}
\textbf{VideoLLaMA2}~\citep{cheng2024videollama} extends the LLaMA backbone with dedicated visual and audio encoders.
\textbf{Phi-4}~\citep{abdin2024phi} is a 14B-parameter model with native tri-modal support.
\textbf{Qwen3o-Instruct} and \textbf{Qwen3o-Thinking}~\citep{xu2025qwen3omnitechnicalreport} share a single architecture but differ at inference: Qwen3o-Thinking emits a reasoning trace inside \texttt{<think>}\dots\texttt{</think>} tags, whereas Qwen3o-Instruct produces only the final answer.
\textbf{MiniCPM-o 2.6}~\citep{yao2024minicpm} is a lightweight omni-modal model optimized for on-device deployment.
\textbf{Uni-MoE-2}~\citep{li2025unimoe20omniscalinglanguagecentricomnimodal} employs a mixture-of-experts architecture for modality fusion.
\textbf{VITA-1.5}~\citep{fu2025vita15gpt4olevelrealtime} is a video--audio--language model with a dedicated audio encoder.

\paragraph{Proprietary models.}
\textbf{Gemini~2.0} and \textbf{Gemini~2.5}~\citep{team2023gemini} are accessed via the Gemini API. \textbf{GPT-4o-mini}~\citep{openai2024gpt4o} is evaluated via a two-stage pipeline: audio is first processed by GPT-4o-mini Audio using the prompt \textit{``Describe this audio''} to produce a textual description, which replaces the raw audio and is passed alongside the video and text inputs to GPT-4o-mini for final answer generation, as it lacks native audio support \footnote{\label{fn:gpt4o}Since GPT-4o-mini transcribes audio 
to text via GPT-4o-mini Audio, its results are not directly 
comparable to native tri-modal models; marked with $\dag$ in 
all tables.}.

\subsection{Evaluation Protocol}
\label{sec:eval-protocol}

\paragraph{Data splits and conditions.}
The benchmark has a \textbf{synthetic} and a \textbf{real} split.
A \emph{corruption vector} $\mathbf{c}=(c_v,c_a,c_t)\in\{0,1\}^3$ indicates which modalities have been replaced ($c_m{=}1$) or left intact ($c_m{=}0$); the \emph{corruption level} $k=\sum_m c_m$ counts the corrupted modalities.
The congruent condition $\mathbf{c}=(0,0,0)$ denotes that each modality in the sample represents the same anchor.

\paragraph{Prompting.}
Each model is tested under \textbf{zero-shot} (ZS) and \textbf{chain-of-thought} (CoT) prompting ($2\text{ splits}\times 2\text{ prompts} = 4$ settings per model).

\paragraph{Answer parsing.}
We extract the selected option (A--E) from model response.
Unparseable outputs are marked \emph{invalid} and excluded from accuracy calculation. 

\paragraph{Ground truth and abstention.}
Ground truth is established through a structured human evidence protocol (Appendix~\ref{app:annotation}).
For each instance, two independent annotators first identify what each modality depicts in isolation, then determine which answer option - if any has strictly greater modality support than all others.
If one option has clear majority evidence support, that option is $y^*$; if evidence is tied, contradictory, or no modality is clear, the annotators select the abstention option as $y^*$.
A third annotator resolves all disagreements by majority vote.
This evidence-grounded labeling reflects what a careful human observer could determine from the input, motivated by findings that MLLMs are severely overconfident under conflicting inputs~\citep{chen2025unveiling,yan2025multimodal} and by selective prediction norms~\citep{geifman2017selective}.

\subsection{Metrics}
\label{sec:metrics}

\paragraph{Accuracy.}
Let $\hat{a}_i$ be the model's prediction and $y_i^*$ the ground truth for instance~$i$.
Per-condition accuracy is $\mathrm{Acc}(\mathbf{c}) = \frac{1}{N_{\mathbf{c}}}\sum_{i=1}^{N_{\mathbf{c}}} \mathbbm{1}[\hat{a}_i = y_i^*]$, where $N_{\mathbf{c}}$ is the number of valid instances under condition~$\mathbf{c}$.
Because $y^*$ encodes the expected target, this metric captures both factual correctness ($k{\leq}1$) and correct abstention ($k{\geq}2$).

\paragraph{Abstention Calibration Error (ACE).}
Let $\mathrm{Abs}_k$ be the model's observed abstention rate at corruption level~$k$, and $\mathrm{Abs}^*_k$ the human-annotated abstention rate at level~$k$, derived from the evidence protocol (Appendix~\ref{app:annotation}):
$\mathrm{ACE} = \frac{1}{4}\sum_{k=0}^{3}\bigl|\mathrm{Abs}_k - \mathrm{Abs}^*_k\bigr|$.
A score of 0 indicates perfect alignment with human abstention behavior; higher scores indicate greater miscalibration.

\paragraph{Confidence calibration.}
For models exposing per-token logits, we compute \textit{restricted softmax}~(RS) over the five answer-option logits~\citep{hendrycks2017baseline}; the per-instance confidence is the maximum of the resulting distribution.
For models providing only scalar log-probabilities, we use the \textit{token probability}~(TP) of the selected answer token as the confidence score.
RS and TP are on different scales and not directly comparable.
We report \textit{mean predictive confidence} $\overline{\mathrm{Conf}}_k = \frac{1}{N_k}\sum_{i} \mathrm{conf}_i$ at each corruption level~$k$, as well as \textit{Expected Calibration Error} (ECE;~\citealp{naeini2015obtaining,guo2017calibration}) and \textit{Risk--Coverage AUC} (RC-AUC;~\citealp{geifman2017selective}), all separated by extraction method.

\paragraph{Modality reliance.}
Additionally, we compute Shapley values~\citep{lundberg2017unified} to attribute accuracy to individual modalities, treating the subset of clean modalities and accuracy as the value function (details in Appendix~\ref{app:shapley}).
We also report \textbf{normalized reliance}
$\Delta_m^{\mathrm{norm}} {=} (\mathrm{Acc}(\mathbf{c}_{000}) {-} \mathrm{Acc}(\mathbf{c}_{m})) / \mathrm{Acc}(\mathbf{c}_{000})$,
capturing the fractional accuracy drop when modality~$m$ alone is corrupted.

\section{Results}
\label{sec:results}

Tables~\ref{tab:accuracy}--\ref{tab:confidence} capture the three central dimensions of model reliability - factual accuracy, abstention behavior, and predictive confidence - across all four settings (zero-shot vs.\ chain-of-thought $\times$ synthetic vs.\ real). Within each table, columns progress from $k{=}0$ to $k{=}3$ left-to-right, making the degradation trajectory immediately visible: how quickly color shifts within a row reveals whether a model adapts to increasing corruption or ignores it.


\begin{table*}[t]
\centering
\renewcommand{\arraystretch}{1.30}
\tiny
\begin{tabular}{
  @{\hspace{2pt}} l @{\hspace{10pt}}
  cccc
  @{\hspace{14pt}}
  cccc
  @{\hspace{5pt}}!{\vrule width 1.4pt}@{\hspace{5pt}}
  cccc
  @{\hspace{14pt}}
  cccc
  @{\hspace{2pt}}
}
\toprule
& \multicolumn{8}{c}{\textbf{Zero-Shot}}
& \multicolumn{8}{c}{\textbf{Chain-of-Thought}} \\[1pt]
\cmidrule(lr){2-9} \cmidrule(l){10-17}
& \multicolumn{4}{c}{\textit{Synthetic}}
& \multicolumn{4}{c}{\textit{Real}}
& \multicolumn{4}{c}{\textit{Synthetic}}
& \multicolumn{4}{c}{\textit{Real}} \\
\cmidrule(lr){2-5} \cmidrule(lr){6-9}
\cmidrule(lr){10-13} \cmidrule(l){14-17}
\textbf{Model}
  & {$k{=}0$} & {$\bar{k}{=}1$} & {$\bar{k}{=}2$} & {$k{=}3$}
  & {$k{=}0$} & {$\bar{k}{=}1$} & {$\bar{k}{=}2$} & {$k{=}3$}
  & {$k{=}0$} & {$\bar{k}{=}1$} & {$\bar{k}{=}2$} & {$k{=}3$}
  & {$k{=}0$} & {$\bar{k}{=}1$} & {$\bar{k}{=}2$} & {$k{=}3$} \\
\midrule
Gemini 2.5   & \gc{96.5}{96}  & \gc{91.0}{91}  & \gc{70.5}{70}  & \gc{60.0}{60}  & \gc{96.8}{97}  & \gc{91.2}{91}  & \gc{69.3}{69}  & \gc{59.4}{59}  & \gc{97.2}{97}  & \gc{94.8}{95}  & \gc{73.8}{74}  & \gc{58.3}{58}  & \gc{97.2}{97}  & \gc{93.4}{93}  & \gc{71.4}{71}  & \gc{53.2}{53} \\
Qwen3o-Think & \gc{93.7}{94}  & \gc{86.7}{87}  & \gc{68.1}{68}  & \gc{61.4}{61}  & \gc{92.2}{92}  & \gc{88.0}{88}  & \gc{68.9}{69}  & \gc{60.0}{60}  & \gc{91.8}{92}  & \gc{88.6}{89}  & \gc{64.5}{64}  & \gc{69.0}{69}  & \gc{92.9}{93}  & \gc{85.1}{85}  & \gc{66.3}{66}  & \gc{71.8}{72} \\
VideoLLaMA2  & \gc{91.0}{91}  & \gc{83.8}{84}  & \gc{68.2}{68}  & \gc{16.1}{16}  & \gc{90.2}{90}  & \gc{82.9}{83}  & \gc{66.1}{66}  & \gc{14.1}{14}  & \gc{91.2}{91}  & \gc{85.4}{85}  & \gc{70.0}{70}  & \gc{14.2}{14}  & \gc{93.4}{93}  & \gc{84.2}{84}  & \gc{67.2}{67}  & \gc{13.2}{13} \\
Qwen3o-Inst  & \gc{91.0}{91}  & \gc{83.9}{84}  & \gc{63.8}{64}  & \gc{65.1}{65}  & \gc{88.2}{88}  & \gc{81.2}{81}  & \gc{63.9}{64}  & \gc{66.7}{67}  & \gc{95.3}{95}  & \gc{88.7}{89}  & \gc{67.9}{68}  & \gc{57.1}{57}  & \gc{92.2}{92}  & \gc{87.5}{88}  & \gc{67.5}{68}  & \gc{57.1}{57} \\
Gemini 2.0   & \gc{89.8}{90}  & \gc{78.2}{78}  & \gc{51.9}{52}  & \gc{81.8}{82}  & \gc{91.5}{92}  & \gc{78.4}{78}  & \gc{51.5}{52}  & \gc{80.1}{80}  & \gc{95.6}{96}  & \gc{91.0}{91}  & \gc{66.3}{66}  & \gc{53.0}{53}  & \gc{94.8}{95}  & \gc{90.4}{90}  & \gc{65.6}{66}  & \gc{58.8}{59} \\
GPT-4o-mini\textsuperscript{\hyperref[fn:gpt4o]{$\dag$}}  & \gc{89.4}{89}  & \gc{80.9}{81}  & \gc{61.2}{61}  & \gc{70.6}{71}  & \gc{86.7}{87}  & \gc{79.7}{80}  & \gc{60.3}{60}  & \gc{65.1}{65}  & \gc{93.7}{94}  & \gc{90.4}{90}  & \gc{76.3}{76}  & \gc{44.3}{44}  & \gc{96.1}{96}  & \gc{91.1}{91}  & \gc{75.6}{76}  & \gc{38.0}{38} \\
Uni-MoE-2    & \gc{88.6}{89}  & \gc{78.2}{78}  & \gc{63.4}{63}  & \gc{43.5}{44}  & \gc{87.8}{88}  & \gc{79.0}{79}  & \gc{61.3}{61}  & \gc{39.2}{39}  & \gc{87.9}{88}  & \gc{75.9}{76}  & \gc{57.2}{57}  & \gc{50.7}{51}  & \gc{84.3}{84}  & \gc{75.4}{75}  & \gc{56.7}{57}  & \gc{53.1}{53} \\
MiniCPM      & \gc{87.5}{88}  & \gc{76.3}{76}  & \gc{54.9}{55}  & \gc{63.9}{64}  & \gc{85.1}{85}  & \gc{73.5}{74}  & \gc{52.2}{52}  & \gc{59.2}{59}  & \gc{91.8}{92}  & \gc{83.9}{84}  & \gc{63.6}{64}  & \gc{41.2}{41}  & \gc{90.2}{90}  & \gc{81.9}{82}  & \gc{61.2}{61}  & \gc{32.9}{33} \\
VITA-1.5     & \gc{87.1}{87}  & \gc{77.3}{77}  & \gc{56.6}{57}  & \gc{44.3}{44}  & \gc{87.1}{87}  & \gc{74.7}{75}  & \gc{53.5}{54}  & \gc{45.5}{46}  & \gc{87.4}{87}  & \gc{78.4}{78}  & \gc{57.9}{58}  & \gc{45.0}{45}  & \gc{87.4}{87}  & \gc{76.9}{77}  & \gc{57.3}{57}  & \gc{42.7}{43} \\
Phi-4        & \gc{85.5}{86}  & \gc{76.1}{76}  & \gc{57.1}{57}  & \gc{19.6}{20}  & \gc{86.7}{87}  & \gc{74.8}{75}  & \gc{56.2}{56}  & \gc{17.3}{17}  & \gc{86.0}{86}  & \gc{74.2}{74}  & \gc{59.6}{60}  & \gc{25.2}{25}  & \gc{86.3}{86}  & \gc{75.7}{76}  & \gc{59.0}{59}  & \gc{21.7}{22} \\
\midrule
\textit{Mean} & \textit{90.0} & \textit{81.2} & \textit{61.6} & \textit{52.6} & \textit{89.2} & \textit{80.3} & \textit{60.3} & \textit{50.7} & \textit{91.8} & \textit{85.1} & \textit{65.7} & \textit{45.8} & \textit{91.5} & \textit{84.2} & \textit{64.8} & \textit{44.2} \\
\bottomrule
\end{tabular}
\caption{\textbf{Accuracy (\%)} across corruption levels.
At $k{\leq}1$, values reflect \emph{factual correctness} (higher = better).
At $k{\geq}2$, values reflect the rate of \emph{correct response} per the human annotation protocol (higher = better).
$\bar{k}{=}1,\!2$ averages over the three single/double-corruption conditions.
Models sorted by $k{=}0$ (ZS, Synthetic). ({\setlength{\fboxsep}{0.7pt}\colorbox{goodcol}{\color{white}\tiny\,green\,}} = better).}
\label{tab:accuracy}
\end{table*}


\begin{table*}[t]
\centering
\setlength{\tabcolsep}{3pt}
\renewcommand{\arraystretch}{1.30}
\tiny
\begin{tabular}{
  @{\hspace{2pt}} l @{\hspace{10pt}}
  cccc
  @{\hspace{14pt}}
  cccc
  @{\hspace{5pt}}!{\vrule width 1.4pt}@{\hspace{5pt}}
  cccc
  @{\hspace{14pt}}
  cccc
  @{\hspace{2pt}}
}
\toprule
& \multicolumn{8}{c}{\textbf{Zero-Shot}}
& \multicolumn{8}{c}{\textbf{Chain-of-Thought}} \\[1pt]
\cmidrule(lr){2-9} \cmidrule(l){10-17}
& \multicolumn{4}{c}{\textit{Synthetic}}
& \multicolumn{4}{c}{\textit{Real}}
& \multicolumn{4}{c}{\textit{Synthetic}}
& \multicolumn{4}{c}{\textit{Real}} \\
\cmidrule(lr){2-5} \cmidrule(lr){6-9}
\cmidrule(lr){10-13} \cmidrule(l){14-17}
\textbf{Model}
  & {$k{=}0$} & {$k{=}1$}
  & {$k{=}2$} & {$k{=}3$}
  & {$k{=}0$} & {$k{=}1$}
  & {$k{=}2$} & {$k{=}3$}
  & {$k{=}0$} & {$k{=}1$}
  & {$k{=}2$} & {$k{=}3$}
  & {$k{=}0$} & {$k{=}1$}
  & {$k{=}2$} & {$k{=}3$} \\
\midrule
\rowcolor{gray!12}
\textit{Human}
  & \textit{0.0} & \textit{0.0} & \textit{1.2} & \textit{97.6}
  & \textit{0.0} & \textit{0.0} & \textit{1.8} & \textit{98.4}
  & \textit{0.0} & \textit{0.0} & \textit{1.2} & \textit{97.6}
  & \textit{0.0} & \textit{0.0} & \textit{1.8} & \textit{98.4} \\
\midrule
Gemini 2.5   & \gc{1.2}{99}   & \gc{5.5}{94}   & \gc{20.0}{80}  & \gc{59.6}{60}  & \gc{1.2}{99}   & \gc{4.6}{95}   & \gc{18.8}{81}  & \gc{58.5}{58}  & \gc{2.0}{98}   & \gc{2.9}{97}   & \gc{17.5}{82}  & \gc{56.2}{56}  & \gc{0.4}{100}  & \gc{3.4}{97}   & \gc{18.2}{82}  & \gc{51.9}{52} \\
Qwen3o-Think & \gc{2.0}{98}   & \gc{6.1}{94}   & \gc{22.6}{77}  & \gc{61.0}{61}  & \gc{3.5}{96}   & \gc{5.9}{94}   & \gc{19.0}{81}  & \gc{58.8}{59}  & \gc{2.4}{98}   & \gc{6.8}{93}   & \gc{28.4}{72}  & \gc{69.0}{69}  & \gc{3.1}{97}   & \gc{10.2}{90}  & \gc{26.7}{73}  & \gc{71.4}{71} \\
Qwen3o-Inst  & \gc{3.9}{96}   & \gc{10.5}{90}  & \gc{28.5}{72}  & \gc{63.9}{64}  & \gc{7.8}{92}   & \gc{13.2}{87}  & \gc{26.8}{73}  & \gc{66.3}{66}  & \gc{2.4}{98}   & \gc{6.9}{93}   & \gc{22.4}{78}  & \gc{57.5}{58}  & \gc{3.9}{96}   & \gc{7.1}{93}   & \gc{22.6}{77}  & \gc{56.7}{57} \\
GPT-4o-mini\textsuperscript{\hyperref[fn:gpt4o]{$\dag$}}  & \gc{5.9}{94}   & \gc{14.5}{86}  & \gc{31.4}{69}  & \gc{70.2}{70}  & \gc{8.2}{92}   & \gc{14.5}{86}  & \gc{30.8}{69}  & \gc{65.1}{65}  & \gc{2.7}{97}   & \gc{3.7}{96}   & \gc{12.8}{87}  & \gc{42.0}{42}  & \gc{0.8}{99}   & \gc{3.5}{96}   & \gc{11.0}{89}  & \gc{37.3}{37} \\
Gemini 2.0   & \gc{7.5}{92}   & \gc{18.2}{82}  & \gc{43.8}{56}  & \gc{81.4}{81}  & \gc{7.3}{93}   & \gc{19.0}{81}  & \gc{43.1}{57}  & \gc{80.9}{81}  & \gc{3.2}{97}   & \gc{5.4}{95}   & \gc{24.2}{76}  & \gc{50.8}{51}  & \gc{3.6}{96}   & \gc{6.2}{94}   & \gc{25.3}{75}  & \gc{58.4}{58} \\
Uni-MoE-2    & \gc{4.3}{96}   & \gc{9.8}{90}   & \gc{19.5}{80}  & \gc{42.7}{43}  & \gc{4.7}{95}   & \gc{9.3}{91}   & \gc{16.3}{84}  & \gc{38.4}{38}  & \gc{5.4}{95}   & \gc{12.7}{87}  & \gc{26.6}{73}  & \gc{49.8}{50}  & \gc{6.6}{93}   & \gc{14.3}{86}  & \gc{25.6}{74}  & \gc{51.9}{52} \\
VideoLLaMA2  & \gc{0.4}{100}  & \gc{2.5}{98}   & \gc{5.0}{95}   & \gc{14.9}{15}  & \gc{1.2}{99}   & \gc{2.0}{98}   & \gc{5.2}{95}   & \gc{13.7}{14}  & \gc{0.9}{99}   & \gc{2.2}{98}   & \gc{4.7}{95}   & \gc{12.7}{13}  & \gc{0.4}{100}  & \gc{2.4}{98}   & \gc{4.6}{95}   & \gc{12.7}{13} \\
MiniCPM      & \gc{8.2}{92}   & \gc{16.1}{84}  & \gc{32.9}{67}  & \gc{63.1}{63}  & \gc{7.5}{92}   & \gc{16.1}{84}  & \gc{32.5}{68}  & \gc{60.0}{60}  & \gc{3.5}{96}   & \gc{6.5}{94}   & \gc{17.8}{82}  & \gc{39.6}{40}  & \gc{2.7}{97}   & \gc{6.7}{93}   & \gc{16.5}{84}  & \gc{32.2}{32} \\
Phi-4        & \gc{2.4}{98}   & \gc{4.6}{95}   & \gc{10.1}{90}  & \gc{18.4}{18}  & \gc{1.6}{98}   & \gc{4.8}{95}   & \gc{9.4}{91}   & \gc{17.6}{18}  & \gc{3.3}{97}   & \gc{6.1}{94}   & \gc{11.4}{89}  & \gc{23.2}{23}  & \gc{3.6}{96}   & \gc{5.9}{94}   & \gc{11.8}{88}  & \gc{22.5}{22} \\
VITA-1.5     & \gc{5.1}{95}   & \gc{12.0}{88}  & \gc{25.9}{74}  & \gc{44.3}{44}  & \gc{4.3}{96}   & \gc{13.9}{86}  & \gc{26.9}{73}  & \gc{45.1}{45}  & \gc{5.1}{95}   & \gc{10.2}{90}  & \gc{23.7}{76}  & \gc{44.6}{45}  & \gc{5.5}{94}   & \gc{11.8}{88}  & \gc{23.4}{77}  & \gc{42.7}{43} \\
\bottomrule
\end{tabular}
\caption{\textbf{Abstention rates (\%)} across corruption levels.
The \textit{Human} row shows evidence-based target rates from our annotation protocol: abstention is near-zero at $k{\leq}2$ (the remaining clean modalities suffice) and near-universal at $k{=}3$.
Cell shading reflects proximity to human targets (\setlength{\fboxsep}{1pt}\colorbox{goodcol}{\color{white}\scriptsize green} = close, \setlength{\fboxsep}{1pt}\colorbox{badcol}{\color{white}\scriptsize red} = far).
ACE measures the average gap between model and human rates across levels
(Appendix~Table~\ref{tab:ace_appendix}).
Models sorted by ZS ACE (Synthetic).}
\label{tab:abstention}
\end{table*}


\begin{table*}[t]
\centering
\renewcommand{\arraystretch}{1.30}
\tiny
\begin{tabular}{
  @{\hspace{2pt}} l @{\hspace{8pt}}
  cccc
  @{\hspace{12pt}}
  cccc
  @{\hspace{5pt}}!{\vrule width 1.4pt}@{\hspace{5pt}}
  cccc
  @{\hspace{12pt}}
  cccc
  @{\hspace{8pt}} c
  @{\hspace{2pt}}
}
\toprule
& \multicolumn{8}{c}{\textbf{Zero-Shot}}
& \multicolumn{8}{c}{\textbf{Chain-of-Thought}} & \\[1pt]
\cmidrule(lr){2-9} \cmidrule(lr){10-17}
& \multicolumn{4}{c}{\textit{Synthetic}}
& \multicolumn{4}{c}{\textit{Real}}
& \multicolumn{4}{c}{\textit{Synthetic}}
& \multicolumn{4}{c}{\textit{Real}} & \\
\cmidrule(lr){2-5} \cmidrule(lr){6-9}
\cmidrule(lr){10-13} \cmidrule(lr){14-17}
\textbf{Model}
  & {$k{=}0$} & {$k{=}1$} & {$k{=}2$} & {$k{=}3$}
  & {$k{=}0$} & {$k{=}1$} & {$k{=}2$} & {$k{=}3$}
  & {$k{=}0$} & {$k{=}1$} & {$k{=}2$} & {$k{=}3$}
  & {$k{=}0$} & {$k{=}1$} & {$k{=}2$} & {$k{=}3$}
  & {\textbf{M.}} \\
\midrule
Gemini 2.5    & \gc{99.7}{99}  & \gc{99.3}{98}  & \gc{98.5}{96}  & \gc{96.9}{92}  & \gc{99.8}{99}  & \gc{99.3}{98}  & \gc{98.5}{96}  & \gc{98.1}{95}  & \gc{99.8}{99}  & \gc{99.6}{99}  & \gc{99.0}{98}  & \gc{98.3}{96}  & \gc{99.5}{99}  & \gc{99.5}{99}  & \gc{99.2}{98}  & \gc{98.8}{97}  & TP \\
GPT-4o-mini\textsuperscript{\hyperref[fn:gpt4o]{$\dag$}}   & \gc{99.1}{98}  & \gc{97.6}{94}  & \gc{96.1}{90}  & \gc{95.8}{89}  & \gc{98.8}{97}  & \gc{97.5}{94}  & \gc{96.3}{91}  & \gc{95.9}{90}  & \gc{99.9}{100} & \gc{99.7}{99}  & \gc{99.5}{99}  & \gc{99.2}{98}  & \gc{99.6}{99}  & \gc{99.7}{99}  & \gc{99.7}{99}  & \gc{99.0}{98}  & TP \\
VITA-1.5      & \gc{93.1}{83}  & \gc{88.1}{70}  & \gc{81.0}{52}  & \gc{75.5}{39}  & \gc{92.1}{80}  & \gc{86.4}{66}  & \gc{80.0}{50}  & \gc{72.2}{31}  & \gc{84.5}{61}  & \gc{78.8}{47}  & \gc{72.3}{31}  & \gc{63.8}{9}   & \gc{79.1}{48}  & \gc{75.0}{38}  & \gc{68.1}{20}  & \gc{60.7}{2}   & TP \\
\midrule
Qwen3o-Think  & \gc{99.0}{98}  & \gc{98.0}{95}  & \gc{96.2}{90}  & \gc{93.6}{84}  & \gc{98.6}{96}  & \gc{98.2}{96}  & \gc{95.2}{88}  & \gc{93.1}{83}  & \gc{100.0}{100}& \gc{100.0}{100}& \gc{100.0}{100}& \gc{100.0}{100}& \gc{100.0}{100}& \gc{100.0}{100}& \gc{100.0}{100}& \gc{100.0}{100}& RS \\
Qwen3o-Inst   & \gc{94.9}{87}  & \gc{91.2}{78}  & \gc{86.8}{67}  & \gc{80.5}{51}  & \gc{95.3}{88}  & \gc{92.1}{80}  & \gc{86.3}{66}  & \gc{80.6}{51}  & \gc{100.0}{100}& \gc{100.0}{100}& \gc{100.0}{100}& \gc{100.0}{100}& \gc{100.0}{100}& \gc{100.0}{100}& \gc{100.0}{100}& \gc{100.0}{100}& RS \\
VideoLLaMA2   & \gc{93.2}{83}  & \gc{89.9}{75}  & \gc{83.1}{58}  & \gc{72.8}{32}  & \gc{94.0}{85}  & \gc{89.9}{75}  & \gc{83.2}{58}  & \gc{70.9}{27}  & \gc{99.9}{100} & \gc{99.9}{100} & \gc{99.7}{99}  & \gc{99.1}{98}  & \gc{100.0}{100}& \gc{99.8}{99}  & \gc{99.6}{99}  & \gc{99.3}{98}  & RS \\
Uni-MoE-2     & \gc{91.8}{80}  & \gc{86.0}{65}  & \gc{79.0}{48}  & \gc{68.1}{20}  & \gc{92.2}{80}  & \gc{86.6}{66}  & \gc{78.4}{46}  & \gc{70.4}{26}  & \gc{97.2}{93}  & \gc{91.6}{79}  & \gc{88.6}{71}  & \gc{89.9}{75}  & \gc{93.9}{85}  & \gc{92.7}{82}  & \gc{85.1}{63}  & \gc{79.4}{49}  & RS \\
Phi-4         & \gc{91.7}{79}  & \gc{86.9}{67}  & \gc{80.2}{51}  & \gc{72.4}{31}  & \gc{91.9}{80}  & \gc{87.5}{69}  & \gc{80.9}{52}  & \gc{72.3}{31}  & \gc{100.0}{100}& \gc{99.7}{99}  & \gc{99.7}{99}  & \gc{99.8}{99}  & \gc{99.8}{99}  & \gc{99.8}{99}  & \gc{99.8}{99}  & \gc{99.9}{100} & RS \\
\bottomrule
\end{tabular}
\caption{\textbf{Mean predictive confidence (\%)} across corruption levels.
Persistent high confidence as $k$ increases indicates overconfidence.
Models sorted by $k{=}0$ confidence (ZS, Synthetic) within each group.
\textbf{M.}: confidence extraction method -
TP = token probability (full vocabulary);
RS = restricted softmax (5 answer options).
TP and RS are on different scales and are not directly comparable.
Gemini~2.0 and MiniCPM omitted (no token-level data available).
Aggregate ECE and RC-AUC are in Appendix~Table~\ref{tab:ece_appendix}. ({\setlength{\fboxsep}{0.7pt}\colorbox{goodcol}{\color{white}\tiny\,green\,}} = higher confidence).}
\label{tab:confidence}
\end{table*}


\subsection{Corruption Degrades Accuracy Sharply and Asymmetrically}
\label{sec:results_accuracy}

Table~\ref{tab:accuracy} shows two clear patterns.
Under clean or single-corruption input ($k{\leq}1$), every model exceeds 85\% at $k{=}0$ and the average $k{=}1$ drop is only 9\,pp, suggesting that models handle the task well when most modalities remain intact.
However, the drop is asymmetric across modalities (Appendix~Table~\ref{tab:percond}): text corruption causes the steepest decline (mean 19.5\,pp), vision a smaller one (6.1\,pp), and audio has minimal effect, indicating that current models rely heavily on textual cues even in an omni-modal setting.

At $k{\geq}2$, where accuracy reflects correct response per the human evidence protocol (factual answer or abstention as appropriate), models split into two groups.
\emph{Abstention-capable} models (Gemini~2.0, Qwen3o variants, Gemini~2.5, MiniCPM, GPT-4o-mini) reach 33--82\% at $k{=}3$, while \emph{abstention-resistant} models (VideoLLaMA2, Phi-4) remain below 20\% under zero-shot and below 26\% under CoT - they almost never abstain, even under full corruption.

CoT prompting presents a trade-off: it improves clean accuracy but can hurt abstention.
The clearest case is Gemini~2.0, where CoT raises $k{=}0$ from 90\% to 96\% yet drops $k{=}3$ abstention from 81\% to 51\%, suggesting that extended reasoning makes models more committed to an answer rather than more cautious.
Results are stable across synthetic and real splits (mean absolute difference under 2\,pp).

\subsection{Calibrated Abstention Remains an Open Problem}
\label{sec:results_abstention}

Table~\ref{tab:abstention} shows that \textit{no model achieves calibrated abstention}.

The best-performing model, Gemini~2.5 (ACE\,${\leq}$\,15.8 under ZS), achieves the closest alignment with human abstention behavior, though it still substantially underabstains at $k{=}2$ and $k{=}3$.
VideoLLaMA2 and Phi-4 show the opposite failure pattern: near-zero false abstention at $k{\leq}1$ but almost no abstention at $k{=}2$ (5--12\%) or $k{=}3$ (13--23\%), yielding ACE values of 22.4 and 23.8 respectively - they respond confidently regardless of how conflicted the evidence is.
No model manages both low false-abstention and high correct-abstention aligned with human judgment.

Two systematic patterns emerge. \textbf{First,} abstention increases monotonically with $k$ for every model and setting, confirming some sensitivity to cross-modal dissonance. However, this sensitivity is miscalibrated in both directions: models abstain too often when evidence is sufficient ($k{\leq}2$, where human annotators almost never abstain) and too rarely when evidence is genuinely conflicting ($k{=}3$, where humans abstain on ${\approx}98\%$ of instances yet no model exceeds 82\%).

\textbf{Second,} CoT \emph{improves} abstention calibration for eight of ten models (Appendix~Table~\ref{tab:ace_appendix}), with only VideoLLaMA2 and Uni-MoE-2 showing slight degradation. At $k{=}2$, where the factual answer is almost always recoverable from the single remaining clean modality, CoT helps models correctly answer rather than over-abstain.

\subsection{Confidence Calibration Under Corruption}
\label{sec:results_confidence}

Table~\ref{tab:confidence} reports mean predictive confidence across corruption levels for eight models with token-level data. The central finding is \textbf{systematic overconfidence}: models remain highly confident even as corruption renders their answers unreliable.

Under zero-shot, confidence declines with $k$ but far too slowly relative to accuracy. At $k{=}2$, mean accuracy drops to approximately 61\% while mean confidence stays at or above 78\% for every model. Gemini~2.5 is the most extreme case: confidence barely moves from 99.7\% to 98.5\% as accuracy drops from 96.5\% to 70.5\%. VITA-1.5 and Uni-MoE-2 show the steepest confidence drops at $k{=}3$ (to 76\% and 68\% respectively), making them the least overconfident despite still being poorly calibrated.

CoT \emph{amplifies} overconfidence. Six of eight models with token-level data show confidence saturated at or near 100\% across all corruption levels under CoT, erasing the gradient visible under ZS entirely. Only VITA-1.5 and Uni-MoE-2 retain meaningful confidence declines under CoT (VITA-1.5: 85$\to$64\% on synthetic; Uni-MoE-2: 94$\to$79\% on real), both showing steeper drops than the remaining six models.

Aggregate ECE and RC-AUC (Appendix~Table~\ref{tab:ece_appendix}) confirm this: ECE values range from 12--26\% under ZS and worsen under CoT, reaching 37\% for Phi-4 and 29\% for Uni-MoE-2.

\subsection{Modality Reliance}
\label{sec:results_reliance}

Two complementary metrics reveal different aspects of how models use each modality.

\textbf{Normalized reliance} (Appendix~Table~\ref{tab:reliance_appendix}) measures how much accuracy drops when a single modality is corrupted while the other two remain clean.
Text corruption causes by far the largest decline (mean $\Delta_T^{\mathrm{norm}}{=}22\%$), vision a moderate one (7\%), and audio nearly zero (${\leq}$1\%).
This shows that models are most fragile to losing text, likely because the remaining modalities cannot compensate for the textual signal. The per-condition breakdown at $k{=}2$ 
(Appendix~Table~\ref{tab:percond}) reinforces this: accuracy 
averages 79.6\% when only text is clean ($C_{110}$), 68.7\% for 
video ($C_{011}$), but just 36.4\% for audio ($C_{101}$).

\textbf{Shapley values} (Appendix~Table~\ref{tab:shapley_appendix}) measure the average marginal contribution of each modality across all possible combinations of clean and corrupted inputs.
Text is dominant across all ten models, accounting for 44--64\% of total Shapley value in models with positive audio contributions, and rising further in models where audio carries a negative Shapley value.
Audio has negative Shapley values in seven of ten models. 
This is driven primarily by the $C_{101}$ condition (only audio 
clean), where accuracy falls \emph{below} the fully corrupted 
baseline ($C_{111}$) - models appear unable to leverage audio as 
a sole clean signal and are instead misled by it. When other 
modalities are also clean, audio contributes negligibly.

Text is the modality models rely on most immediately (high normalized reliance) and contributes most across all coalition combinations (positive Shapley value), while audio is not merely uninformative but actively detrimental in most models.


\paragraph{Open-ended evaluation.}
To assess whether the multiple-choice format inflates performance, we evaluate Qwen3o-Instruct and Qwen3o-Thinking without answer options, using GPT-4.1-mini and GPT-5-mini as judges (96.6\% inter-judge agreement). Open-ended accuracy drops 25\,pp at $k{=}0$ and over 60\,pp at $k{=}2$ relative to MCQ, confirming that answer options substantially scaffold model performance. The modality reliance hierarchy (text $\gg$ video $>$ audio) and the sharp $k{=}2$ collapse are preserved, indicating these patterns reflect genuine model behavior rather than artifacts of the MCQ format. Full results are in Appendix~\ref{sec:open-ended}.

Additionally, per-anchor accuracy and statistical validation results are available in \ref{app:per_anchor} and \ref{app:stats}.

\subsection{Error Analysis}
\label{sec:error_analysis}

We identify three recurring error types across model failures (examples in Appendix~\ref{app:error_analysis}).

\paragraph{E1: Wrong factual answer.}
Most commonly, models follow corrupted text while ignoring clean video and audio (Figures~\ref{fig:e1_car}).
Even when the anchor is correctly perceived, corrupted text can steer the model's reasoning toward a wrong answer (Figure~\ref{fig:e1_bubblewrap}).
Errors also occur without any corruption, from visual misclassification (Figure~\ref{fig:e1_piano_clean}), faulty reasoning (Figure~\ref{fig:e1_gun}, \ref{fig:e1_train_clean}), etc.

\paragraph{E2: Wrong abstention.}
Models abstain when the evidence supports an answer, either by hallucinating what a clean modality depicts and then reasoning from the false percept (Figures~\ref{fig:e2_tiger},~\ref{fig:e2_horse}), or by lacking the world knowledge needed to answer (Figure~\ref{fig:e2_typewriter}).

\paragraph{E3: Missed abstention.}
When all or most modalities are corrupted, models construct confident narratives from unrelated inputs instead of recognizing that no consistent anchor exists (Figures~\ref{fig:e3_typewriter_qwen},~\ref{fig:e3_cat_minicpm},~\ref{fig:e3_piano_qwen},~\ref{fig:e3_stapler}).

Across all three types, models rarely check whether modalities agree with each other.

\section{Conclusion}

\label{sec:conclusion}

We introduced OMD-Bench, a diagnostic benchmark that tests omni-modal models under controlled cross-modal conflict by systematically corrupting subsets of initially congruent tri-modal inputs. Our evaluation of ten models reveals that current systems rely disproportionately on text - with audio carrying negative Shapley value under corruption in seven of ten models - rarely abstain in proportion to human-annotated evidence, and maintain high confidence throughout.
Chain-of-thought prompting improves abstention alignment with human judgment for most models but amplifies overconfidence in predicted probabilities rather than resolving it. These findings highlight a significant gap between the promise of omni-modal integration and current model behavior, and point to the need for training and inference strategies that balance modality contributions and calibrate uncertainty under conflicting inputs.

\section{Limitations}
 
\paragraph{Anchor Coverage.}
OMD-Bench uses 27 anchors across four semantic categories. While sufficient for factorial analysis ($27 \times \sim$9.4 questions $\times$ 8 conditions = 4,080 instances), this set excludes abstract concepts, culturally specific entities, and perceptually ambiguous objects. We prioritized anchors satisfying strict tri-modal recognizability ($>$93\% human recognition per modality), which necessarily filters out harder or more nuanced cases. Expanding the anchor set would improve ecological validity but risks violating the equal-informativeness assumption that enables controlled attribution.
 
\paragraph{Binary Corruption Model.}
Our protocol replaces entire modality channels with content from a different anchor, producing categorical conflict. Real-world inconsistencies are often graded: partial occlusion, background noise, or temporal misalignment. We chose wholesale replacement because it produces unambiguous semantic conflict with a clear ground truth for abstention, enabling the factorial design. However, this means our findings about modality reliance and abstention calibration may not transfer directly to settings with subtler or continuous degradation.
 
\paragraph{Multiple-Choice Format.}
All instances use five-option MCQ with a fixed abstention option. Our own open-ended evaluation (Appendix~\ref{sec:open-ended}) shows that answer options inflate accuracy by 25 to 60 percentage points, confirming that the MCQ format scaffolds performance. We adopted MCQ to enable standardized evaluation across ten architecturally diverse models with consistent answer parsing, but results should be interpreted as upper bounds on model capability.
 
\paragraph{Text Modality Confounds.}
Text descriptions are human-authored narrative passages that may carry distributional cues (e.g., sentence structure) beyond the anchor identity. The strong text dominance observed in Section~\ref{sec:results_reliance} likely reflects both architectural language bias and the relative richness of textual input. Disentangling these two factors would require controlling for passage informativeness more granularly than our current design permits.
 
\paragraph{Model and Confidence Heterogeneity.}
GPT-4o-mini is evaluated via audio-to-text transcription rather than native tri-modal processing, and two models (Gemini~2.0, MiniCPM) lack token-level logits. Additionally, confidence is extracted via restricted softmax for some models and token probability for others; these scales are not directly comparable. These asymmetries are unavoidable when evaluating a mix of open-weight and proprietary systems but limit the uniformity of cross-model comparisons.
 
\paragraph{Language and Cultural Scope.}
All materials are in English, and knowledge questions draw predominantly on Western cultural references (e.g., Disney films, Chopin competitions). This may advantage models trained on English-centric corpora and limits generalizability to multilingual or culturally diverse deployment contexts.

\section{Ethics Statement}
 
\paragraph{Data and Licensing.}
Real-split videos and audios are collected from copyright-free platforms with permissive licenses, and synthetic media are generated via Sora, Grok, Gemini, and AudioX under their respective terms. Text descriptions are original. No personally identifiable information is present.

\paragraph{Intended Use.}
OMD-Bench is a diagnostic tool for identifying failure modes in omni-modal models. It should not serve as a sole criterion for deployment certification, particularly in safety-critical domains where miscalibrated confidence carries substantially higher stakes.
 
\paragraph{Dual-Use Considerations.}
Our corruption protocol documents model vulnerabilities (text dominance, overconfidence under conflict) that could in principle inform adversarial attacks. However, the corruption patterns are coarse and consistent with previously reported bimodal findings. We believe publicly quantifying these failure modes is necessary for the community to address them.

\bibliography{custom}

\appendix
\clearpage

\section{Appendix}
\label{sec:appendix}

We provide supplementary analyses that expand on the main results in Section~\ref{sec:results}. 

\subsection{Modality Reliance}
\label{app:shapley}

We treat the set of clean modalities as $S\subseteq M{=}\{v,a,t\}$ and accuracy as the value function~$v(S)$, where modalities outside~$S$ are corrupted rather than removed.
The Shapley value for modality~$m$ is:
\begin{equation}
  \phi_m =
  \sum_{S\subseteq M\setminus\{m\}}
  w(S)\,
  \bigl[v(S{\cup}\{m\})-v(S)\bigr],
\end{equation}
where $w(S)={|S|!(|M|{-}|S|{-}1)!}/{|M|!}$.

\begin{table}[h]
\setlength{\tabcolsep}{2pt}
\centering
\small
\begin{tabular}{@{}l rrr rrr@{}}
\toprule
& \multicolumn{3}{c}{\textit{Synthetic}}
& \multicolumn{3}{c}{\textit{Real}} \\
\cmidrule(lr){2-4}\cmidrule(l){5-7}
\textbf{Model} & \textbf{V} & \textbf{A} & \textbf{T}
               & \textbf{V} & \textbf{A} & \textbf{T} \\
\midrule
VideoLLaMA2  & \sv{.28}{37} & \sv{.14}{18} & \sv{\textbf{.33}}{44} & \sv{.27}{35} & \sv{.14}{18} & \sv{\textbf{.35}}{47} \\
Phi-4        & \sv{.24}{37} & \sv{.06}{9}  & \sv{\textbf{.36}}{55} & \sv{.25}{36} & \sv{.05}{7}  & \sv{\textbf{.40}}{57} \\
Uni-MoE-2    & \sv{.14}{31} & \sv{.02}{5}  & \sv{\textbf{.29}}{64} & \sv{.16}{32} & \sv{.02}{4}  & \sv{\textbf{.31}}{64} \\
\midrule
VITA-1.5     & .14 & $-$.04 & \textbf{.32} & .12 & $-$.06 & \textbf{.36} \\
Gemini 2.5   & .18 & $-$.07 & \textbf{.26} & .14 & $-$.04 & \textbf{.27} \\
Qwen3o-Think & .15 & $-$.03 & \textbf{.20} & .14 & $-$.03 & \textbf{.21} \\
Qwen3o-Inst  & .12 & $-$.06 & \textbf{.21} & .06 & $-$.06 & \textbf{.22} \\
MiniCPM      & .12 & $-$.14 & \textbf{.26} & .09 & $-$.12 & \textbf{.29} \\
GPT-4o-mini\textsuperscript{\hyperref[fn:gpt4o]{$\dag$}}  & .02 & $-$.04 & \textbf{.21} & .02 & $-$.03 & \textbf{.22} \\
Gemini 2.0   & .07 & $-$.24 & \textbf{.25} & .09 & $-$.24 & \textbf{.26} \\
\bottomrule
\end{tabular}
\caption{Shapley values (zero-shot).
Gray percentages show each modality's share of total value.
\textbf{Bold}: dominant modality.
Shares omitted when negative values are present.}
\label{tab:shapley_appendix}
\end{table}

\begin{table}[h]
\centering
\small
\begin{tabular}{@{}l ccc ccc@{}}
\toprule
& \multicolumn{3}{c}{\textit{Synthetic}}
& \multicolumn{3}{c}{\textit{Real}} \\
\cmidrule(lr){2-4}\cmidrule(l){5-7}
\textbf{Model} & $\Delta_V$ & $\Delta_A$ & $\Delta_T$
               & $\Delta_V$ & $\Delta_A$ & $\Delta_T$ \\
\midrule
GPT-4o-mini\textsuperscript{\hyperref[fn:gpt4o]{$\dag$}}  & .03 & .04       & .22 & .02 & .01       & .20 \\
Gemini 2.0   & .08 & $-$.00    & .31 & .14 & $-$.00    & .30 \\
Gemini 2.5   & .04 & .00       & .13 & .01 & .02       & .14 \\
MiniCPM      & .10 & $-$.00    & .29 & .07 & $-$.00    & .34 \\
Phi-4        & .09 & .00       & .24 & .11 & .00       & .29 \\
Qwen3o-Inst  & .06 & $-$.00    & .17 & .03 & .00       & .21 \\
Qwen3o-Think & .05 & .03       & .14 & .03 & .00       & .11 \\
Uni-MoE-2    & .08 & .02       & .26 & .07 & $-$.00    & .24 \\
VITA-1.5     & .07 & $-$.01    & .27 & .09 & $-$.01    & .35 \\
VideoLLaMA2  & .08 & .02       & .14 & .06 & .01       & .17 \\
\midrule
\textit{Mean} & .07 & .01      & .22 & .06 & .00       & .24 \\
\bottomrule
\end{tabular}
\caption{Normalized reliance $\Delta_m^{\mathrm{norm}}$ (zero-shot).
Higher = more sensitive to that modality's corruption.}
\label{tab:reliance_appendix}
\end{table}

\subsection{Per-Condition Accuracy}
\label{app:percond}

Table~\ref{tab:percond} reports accuracy for each of the eight corruption conditions under zero-shot prompting on the synthetic split. The $C_{101}$ condition (only audio clean) consistently yields the lowest $k{=}2$ accuracy across all models (mean 36.4\%), while $C_{110}$ (only text clean) yields the highest (mean 79.6\%), reinforcing the text dominance discussed in Section~\ref{sec:results_reliance}.

\begin{table*}[h]
\centering
\small
\begin{tabular}{@{}l cccc ccc c@{}}
\toprule
& & \multicolumn{3}{c}{\textit{$k{=}1$}}
    & \multicolumn{3}{c}{\textit{$k{=}2$}} & \\
\cmidrule(lr){3-5}\cmidrule(lr){6-8}
\textbf{Model} & C000 & C100{\tiny(V)} & C010{\tiny(A)} & C001{\tiny(T)}
               & C110{\tiny(VA)} & C101{\tiny(VT)} & C011{\tiny(AT)} & C111 \\
\midrule
Gemini 2.5   & 96.5 & 92.8 & 96.4 & 83.7 & 90.7 & 37.2 & 83.5 & 60.0 \\
Qwen3o-Think & 93.7 & 88.6 & 91.0 & 80.4 & 80.8 & 44.3 & 79.2 & 61.4 \\
VideoLLaMA2  & 91.0 & 83.9 & 89.4 & 78.0 & 78.8 & 51.4 & 74.5 & 16.1 \\
Qwen3o-Inst  & 91.0 & 85.1 & 91.4 & 75.3 & 79.2 & 41.2 & 71.0 & 65.1 \\
Gemini 2.0   & 89.8 & 82.4 & 90.2 & 62.0 & 80.0 & 10.2 & 65.5 & 81.8 \\
GPT-4o-mini\textsuperscript{\hyperref[fn:gpt4o]{$\dag$}}  & 89.4 & 86.7 & 86.3 & 69.8 & 79.2 & 45.1 & 59.2 & 70.6 \\
Uni-MoE-2    & 88.6 & 82.0 & 87.1 & 65.5 & 78.0 & 46.7 & 65.5 & 43.5 \\
MiniCPM      & 87.5 & 78.8 & 87.8 & 62.4 & 76.9 & 23.1 & 64.7 & 63.9 \\
VITA-1.5     & 87.1 & 80.8 & 87.8 & 63.1 & 78.8 & 31.0 & 60.0 & 44.3 \\
Phi-4        & 85.5 & 78.0 & 85.5 & 64.7 & 73.7 & 34.1 & 63.5 & 19.6 \\
\midrule
\textbf{Average} & 90.0 & 83.9 & 89.3 & 70.5 & 79.6 & 36.4 & 68.7 & 52.6 \\
\bottomrule
\end{tabular}
\caption{Per-condition accuracy (\%, synthetic, zero-shot).
C$xyz$ = corruption vector $(c_v,c_a,c_t)$; 1 = corrupted.}
\label{tab:percond}
\end{table*}

\subsection{CoT Effects}
\label{app:cot}

Table~\ref{tab:cot} summarizes the effect of chain-of-thought prompting on clean accuracy and abstention calibration. CoT improves $k{=}0$ accuracy for eight of ten models, with Gemini~2.0 showing the largest gain (+5.8\,pp). It also reduces ACE for eight models, though the improvements are modest (mean $-$0.8\,pp), indicating that CoT helps models answer rather than over-abstain but does not fundamentally resolve calibration failures.

\begin{table}[h]
\centering
\small
\begin{tabular}{@{}l rr rr@{}}
\toprule
& \multicolumn{2}{c}{\textit{Clean acc.}}
& \multicolumn{2}{c}{\textit{Abstention}} \\
\cmidrule(lr){2-3}\cmidrule(l){4-5}
\textbf{Model} & ZS & $\Delta$ & ACE\textsubscript{ZS} & $\Delta$ \\
\midrule
Gemini 2.0   & 0.898 & +0.058 & 0.211 & $-$0.016 \\
GPT-4o-mini\textsuperscript{\hyperref[fn:gpt4o]{$\dag$}}  & 0.894 & +0.043 & 0.195 & $-$0.011 \\
MiniCPM      & 0.875 & +0.043 & 0.226 & $-$0.015 \\
Qwen3o-Inst  & 0.910 & +0.043 & 0.189 & $-$0.012 \\
Gemini 2.5   & 0.965 & +0.007 & 0.158 & $-$0.002 \\
Phi-4        & 0.855 & +0.005 & 0.238 & $-$0.002 \\
VITA-1.5     & 0.871 & +0.003 & 0.238 & $-$0.010 \\
VideoLLaMA2  & 0.910 & +0.002 & 0.224 & +0.004 \\
Uni-MoE-2    & 0.886 & $-$0.007 & 0.218 & +0.011 \\
Qwen3o-Think & 0.937 & $-$0.019 & 0.165 & $-$0.003 \\
\bottomrule
\end{tabular}
\caption{CoT effect (synthetic).
$\Delta$: CoT $-$ ZS.
Positive $\Delta$C000 = CoT helps accuracy.
Positive $\Delta$ACE = CoT \emph{worsens} calibration.}
\label{tab:cot}
\end{table}

\subsection{Abstention Calibration Error}
\label{app:ace}
Table~\ref{tab:ace_appendix} reports ACE across all four evaluation settings. Gemini~2.5 achieves the lowest ACE in every setting (15.6--16.6\%), while Phi-4 and VITA-1.5 consistently score highest (${\geq}$22.7\%), reflecting their near-total reluctance to abstain even under full corruption. ACE values are stable across synthetic and real splits (mean absolute difference 0.6\,pp), confirming that calibration patterns are not split-dependent.

\begin{table}[h]
\centering
\small
\begin{tabular}{@{}l cc cc@{}}
\toprule
& \multicolumn{2}{c}{\textbf{Zero-Shot}}
& \multicolumn{2}{c}{\textbf{Chain-of-Thought}} \\
\cmidrule(lr){2-3} \cmidrule(l){4-5}
\textbf{Model} & \textit{Syn} & \textit{Real} & \textit{Syn} & \textit{Real} \\
\midrule
Gemini 2.5   & \textbf{15.8} & \textbf{15.7} & \textbf{15.6} & 16.6 \\
Qwen3o-Think & 16.5 & 16.5 & 16.3 & \textbf{16.3} \\
Qwen3o-Inst  & 18.9 & 19.5 & 17.7 & 18.4 \\
GPT-4o-mini\textsuperscript{\hyperref[fn:gpt4o]{$\dag$}}  & 19.5 & 21.3 & 18.4 & 18.7 \\
Gemini 2.0   & 21.1 & 21.5 & 19.5 & 18.6 \\
Uni-MoE-2    & 21.8 & 22.1 & 22.9 & 22.8 \\
VideoLLaMA2  & 22.4 & 22.8 & 22.8 & 22.8 \\
MiniCPM      & 22.6 & 23.2 & 21.2 & 22.6 \\
Phi-4        & 23.8 & 23.7 & 23.5 & 23.8 \\
VITA-1.5     & 23.8 & 24.2 & 22.7 & 23.7 \\
\bottomrule
\end{tabular}
\caption{Abstention Calibration Error (ACE, \%, $\downarrow$) across all four settings.
$\mathrm{ACE} = \frac{1}{4}\sum_{k=0}^{3}|\mathrm{Abs}_k - \mathrm{Abs}^*_k|$,
where $\mathrm{Abs}^*_k$ is the human-annotated abstention rate at corruption level~$k$;
0 = perfectly aligned with human abstention behavior.
Models sorted by ZS Synthetic ACE.}
\label{tab:ace_appendix}
\end{table}

\subsection{Aggregate Calibration Error and Risk--Coverage}
\label{app:ece}

Table~\ref{tab:ece_appendix} reports ECE and RC-AUC pooled across all corruption levels. Under zero-shot, ECE ranges from 12.3\% (Uni-MoE-2) to 23.7\% (GPT-4o-mini) for their respective extraction methods. CoT worsens ECE for most models, with Phi-4 reaching 37.0\% on the real split, consistent with the confidence saturation observed in Table~\ref{tab:confidence}.

\begin{table}[h]
\centering
\setlength{\tabcolsep}{2pt}
\small
\begin{tabular}{@{}l rr rr rr rr@{}}
\toprule
& \multicolumn{4}{c}{\textbf{Zero-Shot}}
& \multicolumn{4}{c}{\textbf{Chain-of-Thought}} \\
\cmidrule(lr){2-5} \cmidrule(l){6-9}
& \multicolumn{2}{c}{\textit{Syn}}
& \multicolumn{2}{c}{\textit{Real}}
& \multicolumn{2}{c}{\textit{Syn}}
& \multicolumn{2}{c}{\textit{Real}} \\
\cmidrule(lr){2-3} \cmidrule(lr){4-5}
\cmidrule(lr){6-7} \cmidrule(l){8-9}
\textbf{Model}
  & ECE & RC & ECE & RC & ECE & RC & ECE & RC \\
\midrule
\multicolumn{9}{@{}l}{\textit{Restricted Softmax (RS)}} \\[2pt]
Qwen3o-Inst  & 13.9 &  8.5 & 15.1 &  8.8 & 22.2 & 21.0 & 23.3 & 19.0 \\
Uni-MoE-2    & 12.3 & 11.0 & 13.7 & 11.1 & 29.3 & 36.3 & 20.3 & 27.2 \\
Qwen3o-Think & 19.5 &  6.7 & 19.1 &  6.6 & 20.4 & 20.9 & 23.4 & 19.2 \\
Phi-4        & 20.1 & 15.1 & 21.6 & 15.6 & 34.6 & 23.8 & 37.0 & 23.8 \\
VideoLLaMA2  & 15.2 & 10.0 & 16.6 & 11.1 & 18.4 &  7.4 & 27.3 & 13.2 \\
\midrule
\multicolumn{9}{@{}l}{\textit{Token Probability (TP)}} \\[2pt]
VITA-1.5     & 17.9 & 13.8 & 18.2 & 15.0 &  8.9 & 16.6 &  6.7 & 17.4 \\
Gemini 2.5   & 17.9 & 11.5 & 18.3 & 12.2 & 16.4 & 10.4 & 18.3 & 10.4 \\
GPT-4o-mini\textsuperscript{\hyperref[fn:gpt4o]{$\dag$}}  & 23.7 &  8.3 & 25.5 &  9.6 & 19.9 &  9.8 & 20.4 &  9.4 \\
\bottomrule
\end{tabular}
\caption{Aggregate ECE (\%, $\downarrow$) and RC-AUC (\%, $\downarrow$) computed over all corruption levels pooled.
Lower = better calibrated.
Models grouped by extraction method.}
\label{tab:ece_appendix}
\end{table}

\subsection{Confidence Method Availability}
\label{app:confmethods}
Table~\ref{tab:methods} summarizes which token-level data each model exposes. Five open-weight models provide full logit scores, enabling restricted softmax (RS) extraction. Three models expose only scalar log-probabilities, requiring token probability (TP) extraction. Gemini~2.0 and MiniCPM were not included due to API limitations, and are therefore excluded from confidence calibration analyses in Table~\ref{tab:confidence}.

\begin{table}[h]
\centering
\small
\begin{tabular}{@{}l cc c@{}}
\toprule
\textbf{Model} & \textbf{Scores} & \textbf{LogP} & \textbf{Method} \\
\midrule
VideoLLaMA2  & \cmark & \cmark & RS \\
Phi-4        & \cmark & \cmark & RS \\
Qwen3o-Inst  & \cmark & \cmark & RS \\
Qwen3o-Think & \cmark & \cmark & RS \\
Uni-MoE-2    & \cmark & \cmark & RS \\
\midrule
Gemini 2.5   & \xmark & \cmark & TP \\
GPT-4o-mini\textsuperscript{\hyperref[fn:gpt4o]{$\dag$}}  & \xmark & \cmark & TP \\
VITA-1.5     & \xmark & \cmark & TP \\
\midrule
Gemini 2.0   & \xmark & \xmark & --- \\
MiniCPM      & \xmark & \xmark & --- \\
\bottomrule
\end{tabular}
\caption{Token-level data availability across models.
RS requires per-step logits; TP requires scalar log-probabilities.}
\label{tab:methods}
\end{table}

\begin{figure*}[t]
\centering
\includegraphics[width=\textwidth]{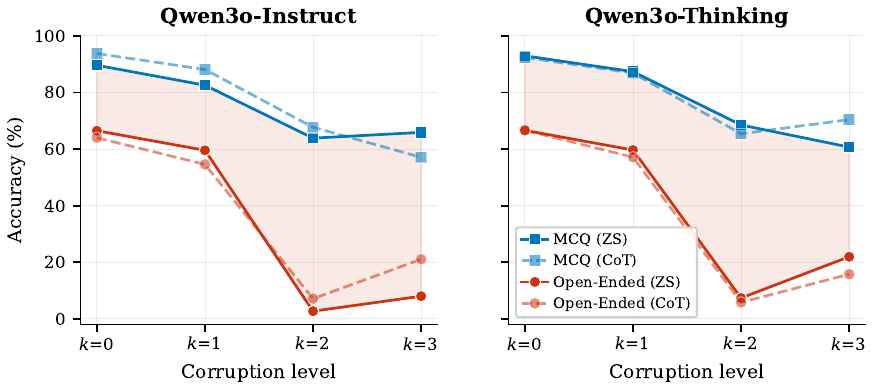}
\caption{MCQ vs.\ open-ended accuracy across corruption levels. Shaded area shows the ZS performance gap. MCQ values are from Table~\ref{tab:accuracy}; open-ended values are averaged across judges and splits.}
\label{fig:oe_mcq_comparison}
\end{figure*}

\begin{figure*}[t]
\centering
\includegraphics[width=\textwidth]{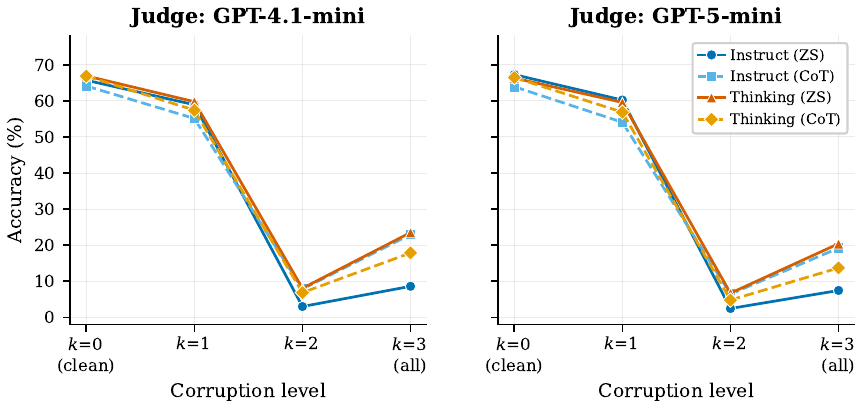}
\caption{Open-ended accuracy by corruption level, split by judge. Each line is a (model, prompt) configuration averaged across both data splits.}
\label{fig:oe_accuracy_by_k}
\end{figure*}

\begin{figure*}[t]
\centering
\includegraphics[width=\textwidth]{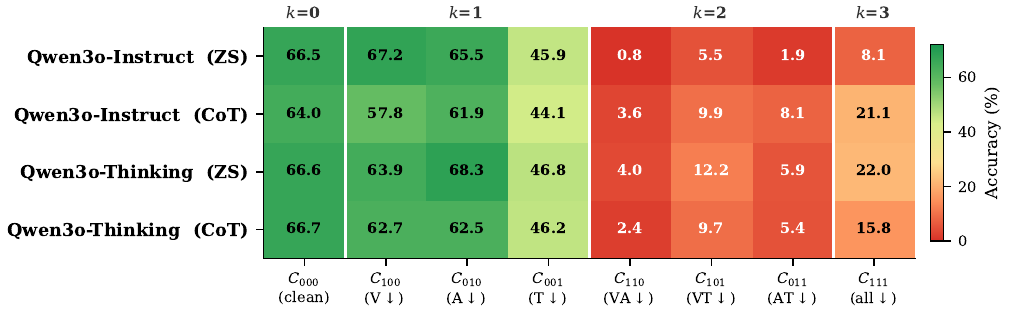}
\caption{Per-condition open-ended accuracy (\%) averaged across judges and splits. Rows show each (model, prompt) configuration. White lines separate corruption levels.}
\label{fig:oe_condition_heatmap}
\end{figure*}

\subsection{Open-Ended Evaluation}
\label{sec:open-ended}
 
We remove all answer options and evaluate Qwen3o-Instruct and Qwen3o-Thinking in an open-ended format.
Models generate free-form responses, which two LLM judges (GPT-4.1-mini and GPT-5-mini) independently score as correct or incorrect against the ground truth.
All~8 corruption conditions, both splits, and both prompting strategies (ZS and CoT) are evaluated, yielding 32,640 judged instances.
 
\paragraph{Inter-judge agreement.}
Table~\ref{tab:oe_judge_agreement} shows that the two judges agree on 96.6\% of instances, with agreement above 94\% at every corruption level and for both models.
Given this high concordance, we report judge-averaged results below.
 
\begin{table}[h]
\centering
\small
\caption{Inter-judge agreement (\%) between GPT-4.1-mini and GPT-5-mini.}
\label{tab:oe_judge_agreement}
\begin{tabular}{lccccc}
\toprule
 & \textbf{Overall} & $k$=0 & $k$=1 & $k$=2 & $k$=3 \\
\midrule
Agreement & 96.6 & 94.6 & 95.5 & 98.4 & 96.8 \\
\bottomrule
\end{tabular}
\end{table}
 
\paragraph{MCQ format inflates accuracy by 25--60~pp.}
Figure~\ref{fig:oe_mcq_comparison} compares MCQ accuracy (from Table~\ref{tab:accuracy}) with open-ended accuracy.
At $k$=0, the gap is $\sim$25~pp (MCQ $\approx$90\% vs.\ open-ended $\approx$66\%), showing that answer options scaffold performance even on clean inputs.
The gap widens to $\sim$60~pp at $k$=2 (MCQ $\approx$66\% vs.\ open-ended $\approx$6\%), where without options models almost entirely fail.
At $k$=3, open-ended accuracy (8--22\%) is below MCQ (57--72\%), as producing appropriate free-text refusals is harder than selecting a pre-defined abstention option.

\paragraph{Accuracy degrades sharply and consistently across judges.}
Figure~\ref{fig:oe_accuracy_by_k} shows degradation trajectories separately for each judge.
Both judges yield nearly identical patterns: accuracy drops moderately from $k$=0 to $k$=1 ($\sim$7~pp), then collapses at $k$=2 (to 3--9\%).
Qwen3o-Thinking slightly outperforms Qwen3o-Instruct under zero-shot, while CoT helps Instruct at $k$=3 but hurts Thinking.

\paragraph{Per-condition breakdown confirms text dominance.}
Figure~\ref{fig:oe_condition_heatmap} shows all~8 conditions.
Text corruption ($C_{001}$) causes the steepest single-modality drop ($\sim$20~pp), while audio corruption ($C_{010}$) has minimal effect ($<$3~pp) - matching the MCQ findings.
At $k$=2, the surviving modality matters: text-only clean ($C_{110}$) yields the lowest accuracy (0.8--4.3\%), indicating that even when text is preserved, models cannot leverage it alone without options.

\subsection{Annotation Procedure and Quality}
\label{app:annotation}

\paragraph{Annotator pool.}
Three annotators, each proficient in English and holding at least a bachelor's degree, 
performed all annotation tasks. Annotators were briefed on the 
task structure, provided with detailed written guidelines 
covering anchor recognition, question formatting, and corruption 
verification, and completed a calibration round on held-out 
examples before annotating the benchmark data.

\paragraph{Question authoring protocol.}
Each annotator independently authored questions for their assigned anchors following a three-step template:
(1)~identify a non-trivial world-knowledge fact about the anchor (e.g., ``rain consists of water droplets''),
(2)~formulate a question requiring both perception of the anchor and retrieval of this fact,
(3)~generate four answer options following the distractor protocol described in \S\ref{sec:distractors}.
Each question was reviewed by at least one additional annotator for factual correctness, unambiguity, and difficulty calibration.

\paragraph{Text description design.}
Text descriptions are narrative passages centered on the anchor 
entity (e.g., a scene involving a dog) and are reviewed to ensure 
they do not contain cues that directly answer any associated 
question. The text dominance observed in \S\ref{sec:results_reliance} 
is consistent with the architectural language bias reported in 
bimodal settings~\citep{chen2024quantifying,zheng2025modality}.

\paragraph{Ground-truth labeling protocol.}
The benchmark questions require world knowledge (e.g., which 
film features a typewriter, which country developed the car) 
that humans may not readily know. We therefore separate 
perception from knowledge in the labeling process.
For each instance, annotators perform two steps:
(i)~identify what each modality depicts by selecting from the 
full list of 27~anchors - a straightforward recognition task 
validated by the tri-modal recognizability criterion in 
\S\ref{sec:anchors}; and
(ii)~look up which answer option corresponds to the identified 
anchor(s) using a provided answer key that maps each anchor to 
its correct option for every question.
If the modalities consistently point to one anchor, the 
corresponding option is the ground truth; if they point to 
different anchors and the evidence is conflicting, annotators 
select abstention.
Annotators thus only need to recognize what each modality 
shows - they never need to know the factual answers themselves.

\paragraph{Inter-annotator agreement.}
All instances were independently labeled by two annotators.
Raw agreement on label correctness was 94.8\% on the real set and 98.0\% on the synthetic set.
Cohen's $\kappa$ for label correctness was 0.625 (substantial) on the real set and 0.796 (substantial) on the synthetic set.
Items where the two annotators disagreed were sent to a third tie-breaking annotator (5.2\%, on the real set; 2.0\%, on the synthetic set), with the final label determined by majority vote.

\paragraph{Corruption verification.}
For each corruption condition, annotators verified that:
(i)~the corrupted modality clearly depicted the replacement anchor $\alpha'$,
(ii)~the corruption created genuine semantic conflict (not merely degraded perceptual quality), and
(iii)~the replacement was perceptually clear in both synthetic and real versions.


\subsection{Corruption Conditions}
\label{app:conditions}

Table~\ref{tab:conditions} lists all eight corruption conditions.
The corruption level $k = \sum_m c_m$ counts the number of replaced modalities.

\begin{table}[h]
  \centering
  \tiny
  \begin{tabular}{lccc cl}
    \toprule
    \textbf{Condition} & $c_v$ & $c_a$ & $c_t$ & $k$ & \textbf{Description} \\
    \midrule
    $C_{000}$ & 0 & 0 & 0 & 0 & Fully congruent (baseline) \\
    $C_{100}$ & 1 & 0 & 0 & 1 & Video corrupted \\
    $C_{010}$ & 0 & 1 & 0 & 1 & Audio corrupted \\
    $C_{001}$ & 0 & 0 & 1 & 1 & Text corrupted \\
    $C_{110}$ & 1 & 1 & 0 & 2 & Video + audio corrupted \\
    $C_{101}$ & 1 & 0 & 1 & 2 & Video + text corrupted \\
    $C_{011}$ & 0 & 1 & 1 & 2 & Audio + text corrupted \\
    $C_{111}$ & 1 & 1 & 1 & 3 & Fully incongruent \\
    \bottomrule
  \end{tabular}
  \caption{Complete corruption conditions.}
  \label{tab:conditions}
\end{table}

\subsection{Data Sources}
\label{app:data_sources}

For the real split, videos are sourced from \href{https://www.pexels.com}{Pexels} and audio from \href{https://freesound.org}{Freesound} and \href{https://pixabay.com/sound-effects/}{Pixabay Sound Effects}. For the synthetic split, videos are generated using Sora, Grok, and Gemini, and audio is synthesized using AudioX \citep{tian2025audiox}. Text descriptions are written by annotators for both splits.

\begin{table}[t]
\centering
\setlength{\tabcolsep}{4.5pt}
\renewcommand{\arraystretch}{1.18}
\small
\begin{tabular}{cl cc}
\toprule
\multirow{2}{*}{\makecell{\textbf{Level}\\\textbf{($k$)}}}
  & \multirow{2}{*}{\textbf{Corrupted Modalities}}
  & \multicolumn{2}{c}{\textbf{Abstention (\%)}} \\
\cmidrule(lr){3-4}
 & & \textbf{Real} & \textbf{Synthetic} \\
\midrule
\multirow{1}{*}{$k{=}0$}
  & None (baseline)
  & \abscell{abs0}{0.00}
  & \abscell{abs0}{0.00} \\
\midrule
\multirow{3}{*}{$k{=}1$}
  & Text only
  & \abscell{abs0}{0.00}
  & \abscell{abs0}{0.00} \\
  & Audio only
  & \abscell{abs0}{0.00}
  & \abscell{abs0}{0.00} \\
  & Visual only
  & \abscell{abs0}{0.00}
  & \abscell{abs0}{0.00} \\
\midrule
\multirow{3}{*}{$k{=}2$}
  & Audio + Text
  & \abscell{abs1}{0.78}
  & \abscell{abs0}{0.39} \\
  & Text + Visual
  & \abscell{abs2}{3.92}
  & \abscell{abs1}{2.75} \\
  & Audio + Visual
  & \abscell{abs1}{0.78}
  & \abscell{abs0}{0.39} \\
\midrule
$k{=}3$
  & Audio + Text + Visual
  & \abscell{abs98}{\textcolor{white}{98.43}}
  & \abscell{abs97}{\textcolor{white}{97.65}} \\
\bottomrule
\end{tabular}

\caption{
  Human-annotated abstention distribution across corruption levels ($k$) and splits.
  Abstention is near-zero for $k \leq 1$, marginal for $k{=}2$ except when text and visual are jointly corrupted, and near-universal (${\approx}98\%$) under full tri-modal corruption ($k{=}3$).
  Cell shading reflects abstention magnitude: \colorbox{abs0}{\phantom{x}} low \textrightarrow{} \colorbox{abs98}{\textcolor{white}{\phantom{x}}} high.
}
\label{tab:abstention-distribution}
\end{table}

\subsection{Statistical Validation}
\label{app:stats}
 
To confirm that the patterns reported in \S\ref{sec:results} are not artifacts of sample size or anchor composition, we conduct four families of statistical tests over the 255 seed questions $\times$ 8 conditions $\times$ 10 models.
All bootstrap intervals use 10{,}000 resamples; all multiple-comparison corrections use Benjamini--Hochberg FDR at $\alpha{=}0.05$.
 
\paragraph{Bootstrap confidence intervals.}
We compute 95\% bootstrap CIs for all per-condition accuracy values from Table~\ref{tab:accuracy}.
The mean CI width across all model--condition cells is 9.6\,pp, narrow enough that the main-text accuracy differences (Table~\ref{tab:accuracy}) are reliable.
Notably, the $C_{101}$ condition (only audio clean) produces the widest CIs for every model, reflecting the high per-anchor variance when audio is the sole remaining signal.
 
\paragraph{Modality reliance is statistically significant.}
We use per-anchor Wilcoxon signed-rank tests to validate the modality hierarchy from \S\ref{sec:results_reliance}.
For each model, we compute the accuracy drop caused by corrupting each modality (relative to $C_{000}$) at the anchor level and test whether drops differ across modality pairs.
Table~\ref{tab:wilcoxon_modality} shows that the text--audio gap is significant ($p<0.05$) for all ten models, confirming that text corruption is universally more damaging than audio corruption.
The text--video gap reaches significance for six of ten models;
the four non-significant cases (VideoLLaMA2, Phi-4, Qwen3o-Instruct, Qwen3o-Thinking) are models where both video and text carry substantial Shapley value (Table~\ref{tab:shapley_appendix}), narrowing the per-anchor gap below detectable levels at $n{=}27$ anchors.
 
\paragraph{Model ranking significance (Friedman test).}
Treating each anchor as a block and each model's accuracy as the observation, a Friedman test rejects the null hypothesis of equal model performance at every corruption level: $p{=}3.5{\times}10^{-6}$ at $k{=}0$, $p{=}4.0{\times}10^{-11}$ at $k{=}1$, $p{=}4.4{\times}10^{-12}$ at $k{=}2$, and $p{=}2.4{\times}10^{-25}$ at $k{=}3$ (all synthetic, zero-shot, df${=}9$).
The monotonically decreasing $p$-values show that model differences \emph{amplify} under corruption: the gap between robust and fragile models widens as more modalities are replaced.
Among the $\binom{10}{2}{=}45$ model pairs at $C_{000}$, McNemar's test (with BH correction) identifies 12 as significantly different, rising to 30{+} pairs at $k{\geq}2$.
 
\paragraph{Chain-of-thought effect.}
Paired permutation tests on matched instances confirm that CoT significantly improves overall accuracy for 13 of 20 model--split combinations ($p{<}0.05$), with the largest gains for Gemini~2.0 ($+7.5$\,pp) and GPT-4o-mini ($+6.5$\,pp).
The two exceptions where ZS significantly outperforms CoT are both Uni-MoE-2 ($+2.8$\,pp syn, $+1.9$\,pp real).
VideoLLaMA2 and Qwen3o-Thinking show no significant difference in either direction.

 
\begin{table}[t]
\centering
\small
\setlength{\tabcolsep}{3.5pt}
\begin{tabular}{@{}l c l @{\hspace{10pt}} c l@{}}
\toprule
& \multicolumn{2}{c}{\textbf{Text vs.\ Video}} & \multicolumn{2}{c}{\textbf{Text vs.\ Audio}} \\
\cmidrule(lr){2-3} \cmidrule(l){4-5}
\textbf{Model} & {$\bar\Delta$\,(pp)} & {$p$} & {$\bar\Delta$\,(pp)} & {$p$} \\
\midrule
Gemini 2.0    & $+$17.5 & .006{**}       & $+$28.8 & ${<}.001${***} \\
Gemini 2.5    & $+$6.5  & .021{*}        & $+$8.5  & .006{**} \\
GPT-4o-mini\textsuperscript{$\dag$} & $+$15.9 & .001{**} & $+$8.6 & .015{*} \\
MiniCPM       & $+$10.7 & .026{*}        & $+$20.8 & ${<}.001${***} \\
Uni-MoE-2     & $+$22.3 & ${<}.001${***} & $+$27.7 & ${<}.001${***} \\
VITA-1.5      & $+$21.9 & .002{**}       & $+$25.8 & ${<}.001${***} \\
\midrule
VideoLLaMA2   & $+$2.7  & .118           & $+$9.6  & .004{**} \\
Phi-4         & $+$4.0  & .223           & $+$20.1 & ${<}.001${***} \\
Qwen3o-Inst   & $+$5.5  & .177           & $+$11.5 & .002{**} \\
Qwen3o-Think  & $+$6.2  & .073           & $+$7.3  & .018{*} \\
\bottomrule
\end{tabular}
\caption{Wilcoxon signed-rank tests for modality reliance
(synthetic, zero-shot).
$\bar\Delta$: mean per-anchor accuracy drop difference
(positive = text corruption hurts more).
Text--audio significance holds for all ten models;
text--video for six.
$^{*}p{<}.05$; $^{**}p{<}.01$; $^{***}p{<}.001$.}
\label{tab:wilcoxon_modality}
\end{table}
 
 
\begin{table}[t]
\centering
\small
\begin{tabular}{@{}l cc cc@{}}
\toprule
& \multicolumn{2}{c}{\textbf{Zero-Shot}} & \multicolumn{2}{c}{\textbf{Chain-of-Thought}} \\
\cmidrule(lr){2-3} \cmidrule(l){4-5}
\textbf{Model} & {ACE} & {95\% CI} & {ACE} & {95\% CI} \\
\midrule
Gemini 2.5    & \textbf{15.9} & [14.1, 17.8] & \textbf{15.6} & [13.9, 17.4] \\
Qwen3o-Think  & 16.5 & [14.7, 18.3] & 16.2 & [14.4, 18.0] \\
Qwen3o-Inst   & 18.8 & [17.0, 20.7] & 17.7 & [15.8, 19.5] \\
GPT-4o-mini\textsuperscript{$\dag$}   & 19.5 & [17.6, 21.4] & 18.4 & [16.7, 20.1] \\
Gemini 2.0    & 21.1 & [19.3, 22.9] & 19.6 & [17.7, 21.5] \\
Uni-MoE-2     & 21.8 & [19.9, 23.7] & 22.8 & [20.8, 24.9] \\
VideoLLaMA2   & 22.3 & [21.1, 23.5] & 22.9 & [21.6, 24.1] \\
MiniCPM       & 22.6 & [20.6, 24.6] & 21.2 & [19.4, 22.9] \\
Phi-4         & 23.7 & [22.3, 25.2] & 23.5 & [21.9, 25.1] \\
VITA-1.5      & 23.8 & [21.9, 25.8] & 22.7 & [20.8, 24.6] \\
\bottomrule
\end{tabular}
\caption{Abstention Calibration Error (ACE, \%, $\downarrow$)
with 95\% bootstrap CIs (synthetic split, 10{,}000 resamples).
CIs confirm that the top two models (Gemini~2.5, Qwen3o-Think)
are separated from the bottom cluster
(VideoLLaMA2, Phi-4, VITA-1.5) by non-overlapping intervals.
Models sorted by ZS ACE.}
\label{tab:ace_ci}
\end{table}

\subsection{Per-Anchor Accuracy}
\label{app:per_anchor}
 
Table~\ref{tab:per_anchor} reports per-anchor accuracy at $k{=}0$ for Gemini~2.5 (mean 97.2\%, $\sigma{=}6.3$\,pp).
Of the 27 anchors, 22 are solved perfectly and only four fall below 90\%, all involving world-knowledge questions that are difficult independently of perception.
No anchor drops below 80\%, confirming that the benchmark does not contain perceptually unrecognizable items.
 
\begin{table}[h]
\centering
\small
\begin{tabular}{@{}lr@{\hspace{18pt}}lr@{}}
\toprule
\textbf{Anchor} & \textbf{Acc.} & \textbf{Anchor} & \textbf{Acc.} \\
\midrule
bubble\_wrap\_pop & 80.0  & kettle\_whistle  & 100.0 \\
kid\_laughing     & 80.0  & kid\_crying      & 100.0 \\
train             & 84.2  & match            & 100.0 \\
clock             & 85.0  & opera            & 100.0 \\
piano             & 95.0  & popcorn\_pop     & 100.0 \\
bee               & 100.0 & rain             & 100.0 \\
bells             & 100.0 & shotgun          & 100.0 \\
bowling           & 100.0 & stapler          & 100.0 \\
car               & 100.0 & t\_rex           & 100.0 \\
cat               & 100.0 & tiger            & 100.0 \\
couple\_yelling   & 100.0 & typewriter       & 100.0 \\
dog               & 100.0 & water\_running   & 100.0 \\
gun               & 100.0 & zipper           & 100.0 \\
horse             & 100.0 &                  &       \\
\bottomrule
\end{tabular}
\caption{Per-anchor accuracy (\%) at $k{=}0$ for Gemini~2.5 (synthetic, zero-shot).
22 of 27 anchors are solved perfectly; the five below 100\% mainly reflect harder knowledge questions.}
\label{tab:per_anchor}
\end{table}

\clearpage


\begin{figure*}[t]
\subsection{Error Analysis}
\label{app:error_analysis}
\subsubsection*{E1: Wrong Factual Answer}
\qualboxbegin{E1H}{E1Light}{E1\ \ Wrong Answer}%
  {$C_{000}$\enspace($k{=}0$, all clean)\enspace Anchor:\ \textit{piano}}

\metabadge{Model: Gemini~2.5}\;
\metabadge{Prompt: zero-shot}\;
\metabadge{Split: real}\;
\metabadge{True anchor: piano}

\vspace{6pt}

\modalitygrid%
  {\cleanpill}%
  {\includegraphics[width=\linewidth]{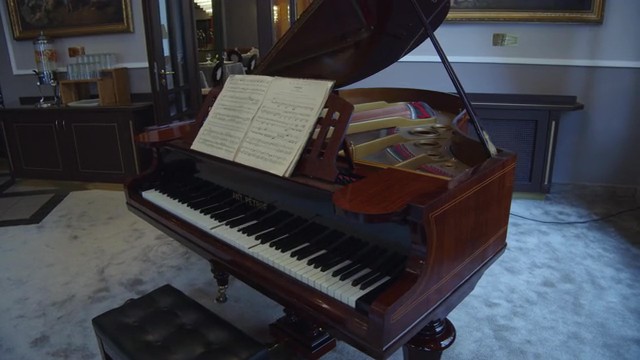}}%
  {\cleanpill}%
  {Piano music\\[2pt]%
   {\fontsize{6.5}{8}\selectfont\textit{Clean - corresponds to true anchor `piano'}}}%
  {\cleanpill}%
  {\textit{``\ldots prestigious international competition returned to Warsaw [\ldots]%
    Polish composer whose nocturnes, \'etudes, and ballades\ldots''}\\[2pt]%
   {\fontsize{6.5}{8}\selectfont Passage describes Chopin piano competition - clean, true anchor}}%
  {Piano being played}%

\vspace{6pt}
{\fontsize{9}{11}\selectfont\textbf{Q:} The frame of the instrument present across the content is constructed from which material?}

\vspace{4pt}
\optionblock{%
  \optPred{A}{Laminated maple}\\[3pt]%
  \optAbs{B}{I cannot answer}\\[3pt]%
  \optOther{C}{Carbon fiber}\\[3pt]%
  \optOther{D}{Brass alloy}\\[3pt]%
  \optGT{E}{Cast iron}\enspace\textcolor{GTGreen}{\qcmark~\textbf{\fontsize{8}{9}\selectfont human GT}}}

\vspace{5pt}
\begin{tcolorbox}[obox]
  {\fontsize{7}{8.5}\selectfont\textbf{Model output (Gemini~2.5, zero-shot):}}\vspace{3pt}\par
  {\fontsize{8}{10}\selectfont\textit{`The instrument shown in the video is a digital piano [\ldots]
    Cast iron is used for the harp/plate inside an acoustic piano to withstand string tension,
    but not for the outer frame or casing of either an acoustic or digital piano [\ldots]
    Based on the visual evidence of the digital piano's dark, wood-like casing,
    Laminated maple' is the most appropriate material.''
    \hfill$\Rightarrow$\,\textbf{predicts A}}}
\end{tcolorbox}

\vspace{4pt}
\begin{tcolorbox}[fbox2]
  {\fontsize{8.5}{10.5}\selectfont\textbf{\textcolor{PredRed}{Why it fails:}}\ All three
  modalities are clean. The model misclassifies the video as showing a \emph{digital} piano
  and uses that percept to explicitly dismiss cast iron - correctly noting it belongs to
  acoustic instruments - despite the text unambiguously describing an acoustic instrument
  at a Chopin competition in Warsaw.
  \textbf{Visual subtype misclassification overrides correct text-grounded reasoning even
  with zero corruption: the model cites the right fact about cast iron, then discards it
  on the basis of a wrong visual identification.}}
\end{tcolorbox}

\qualboxend
\caption{E1: Wrong factual answer - piano ($k{=}0$, Gemini~2.5, real).}
\label{fig:e1_piano_clean}
\end{figure*}

\begin{figure*}[t]
\qualboxbegin{E1H}{E1Light}{E1\ \ Wrong Answer}%
  {$C_{001}$\enspace($k{=}1$, text corrupted)\enspace Anchor:\ \textit{bubble\_wrap\_pop}}

\metabadge{Model: MiniCPM}\;
\metabadge{Prompt: CoT}\;
\metabadge{Split: synthetic}\;
\metabadge{True anchor: bubble\_wrap\_pop}

\vspace{6pt}

\modalitygrid%
  {\cleanpill}%
  {\includegraphics[width=\linewidth]{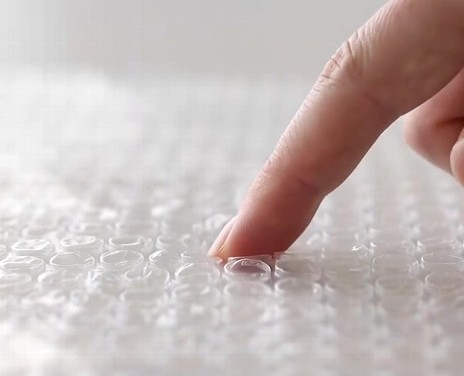}}%
  {\cleanpill}%
  {Bubble wrap popping\\[2pt]%
   {\fontsize{6.5}{8}\selectfont\textit{True anchor - matches bubble\_wrap\_pop'}}}%
  {\corruptpill}%
  {\textit{``\ldots the delivery room fell silent [\ldots] a sharp, wailing shriek as
    the tiny lungs filled with air for the first time [\ldots]
    that unmistakable sound, though piercing, was exactly what everyone needed
    to hear\ldots''}\\[2pt]%
   {\fontsize{6.5}{8}\selectfont Passage describes a newborn delivery room - not bubble\_wrap\_pop'}}%
  {Bubble wrap being popped}%

\vspace{6pt}
{\fontsize{9}{11}\selectfont\textbf{Q:} The material present across the content was originally invented for which purpose?}

\vspace{4pt}
\optionblock{%
  \optOther{A}{I cannot answer}\\[3pt]%
  \optPred{B}{Sterile surgical padding}\enspace$\Leftarrow$\textbf{\textcolor{PredRed}{\fontsize{8}{9}\selectfont predicted}}\\[3pt]%
  \optOther{C}{Greenhouse insulation}\\[3pt]%
  \optGT{D}{Textured wallpaper}\enspace\textcolor{GTGreen}{\qcmark~\textbf{\fontsize{8}{9}\selectfont human GT}}\\[3pt]%
  \optOther{E}{Flotation devices}}

\vspace{5pt}
\begin{tcolorbox}[obox]
  {\fontsize{7}{8.5}\selectfont\textbf{Model output (MiniCPM, CoT):}}\vspace{3pt}\par
  {\fontsize{8}{10}\selectfont\textit{``The image shows a close-up of a hand pressing down on a surface
    with a bubble-like texture [\ldots] given the context of a delivery room and
    the mention of a newborn, it is likely that the bubble wrap is being used to
    protect delicate items, possibly medical equipment or supplies [\ldots]
    the most plausible original purpose of bubble wrap aligns with option B
    (sterile surgical padding) in a broader sense.''
    \hfill$\Rightarrow$\,\textbf{predicts B}}}
\end{tcolorbox}

\vspace{4pt}
\begin{tcolorbox}[fbox2]
  {\fontsize{8.5}{10.5}\selectfont\textbf{\textcolor{PredRed}{Why it fails:}}\ The model
  correctly identifies bubble wrap from the clean video and audio, demonstrating that
  the anchor is perceived accurately. However, the corrupted delivery-room text
  immediately recontextualises the material: the model reinterprets the clean percept
  through a medical frame and selects ``sterile surgical padding'' as the original
  invention purpose. The true answer - textured wallpaper, the failed product concept
  by bubble wrap's inventors in 1957 - is entirely accessible from world knowledge
  once the anchor is identified. \textbf{A single corrupted text channel does not
  change what the model sees, but changes what the model infers from what it sees.}}
\end{tcolorbox}

\qualboxend
\caption{E1: Wrong factual answer - bubble wrap ($k{=}1$, MiniCPM, synthetic).}
\label{fig:e1_bubblewrap}
\end{figure*}

\bigskip

\begin{figure*}[t]
\qualboxbegin{E1H}{E1Light}{E1\ \ Wrong Answer}%
  {$C_{001}$\enspace($k{=}1$, text corrupted)\enspace Anchor:\ \textit{gun}}

\metabadge{Model: GPT-4o-Mini}\;
\metabadge{Prompt: CoT}\;
\metabadge{Split: synthetic}\;
\metabadge{True anchor: gun}

\vspace{6pt}

\modalitygrid%
  {\cleanpill}%
  {\includegraphics[width=\linewidth]{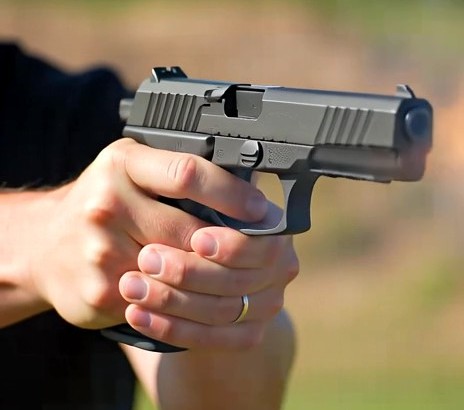}}%
  {\cleanpill}%
  {Gunshots firing\\[2pt]%
   {\fontsize{6.5}{8}\selectfont\textit{True anchor - matches gun'}}}%
  {\corruptpill}%
  {\textit{``\ldots ancient temple walls covered in hieroglyphics depicting the sacred
    animal [\ldots] priests had mummified thousands of them as offerings to Bastet [\ldots]
    killing one, even accidentally, was punishable by death\ldots''}\\[2pt]%
   {\fontsize{6.5}{8}\selectfont Passage describes Egyptian cat worship - not gun'}}%
  {Gun being fired}%

\vspace{6pt}
{\fontsize{9}{11}\selectfont\textbf{Q:} The item present across the content is traditionally used in which ceremonial context?}

\vspace{4pt}
\optionblock{%
  \optOther{A}{Nautical christening rituals}\\[3pt]%
  \optPred{B}{Military funerals and state honors}\enspace\ldots\enspace\textcolor{GTGreen}{\qcmark~\textbf{\fontsize{8}{9}\selectfont human GT}}\\[3pt]%
  \optOther{C}{I cannot answer}\\[3pt]%
  \optPred{D}{Religious ordination rites}\enspace$\Leftarrow$\textbf{\textcolor{PredRed}{\fontsize{8}{9}\selectfont predicted}}\\[3pt]%
  \optOther{E}{Academic convocation ceremonies}}

\vspace{5pt}
\begin{tcolorbox}[obox]
  {\fontsize{7}{8.5}\selectfont\textbf{Model output (GPT-4o-Mini, CoT):}}\vspace{3pt}\par
  {\fontsize{8}{10}\selectfont\textit{``The description of ancient temple walls, hieroglyphics,
    and the reverence for animals in relation to the goddess Bastet indicates a strong
    connection to religious practices in ancient Egypt [\ldots]
    the most fitting option that aligns with the ceremonial context of the ancient
    Egyptian practices described is D.''
    \hfill$\Rightarrow$\,\textbf{predicts D}}}
\end{tcolorbox}

\vspace{4pt}
\begin{tcolorbox}[fbox2]
  {\fontsize{8.5}{10.5}\selectfont\textbf{\textcolor{PredRed}{Why it fails:}}\ Both
  clean modalities unambiguously identify the anchor as a gun: the video shows a firearm
  being discharged and the audio contains explicit gunshot sounds. The model's own
  reasoning acknowledges ``the audio description mentions a loud, sharp sound resembling
  a gunshot,'' yet it immediately pivots entirely to the corrupted text and selects an
  answer solely on the basis of the Egyptian religious passage.
  \textbf{The correct answer - military funerals and the 21-gun salute - is
  directly licensed by the two clean channels the model chose to discard.
  A single corrupted text channel fully overwrites two clean percepts.}}
\end{tcolorbox}

\qualboxend
\caption{E1: Wrong factual answer - gun ($k{=}1$, GPT-4o-Mini, synthetic).}
\label{fig:e1_gun}
\end{figure*}

\bigskip
\begin{figure*}[t]
\qualboxbegin{E1H}{E1Light}{E1\ \ Wrong Answer}%
  {$C_{000}$\enspace($k{=}0$, all clean)\enspace Anchor:\ \textit{train}}
\metabadge{Model: MiniCPM}\;
\metabadge{Prompt: CoT}\;
\metabadge{Split: synthetic}\;
\metabadge{True anchor: train}
\vspace{6pt}
\modalitygrid%
  {\cleanpill}%
  {\includegraphics[width=\linewidth]{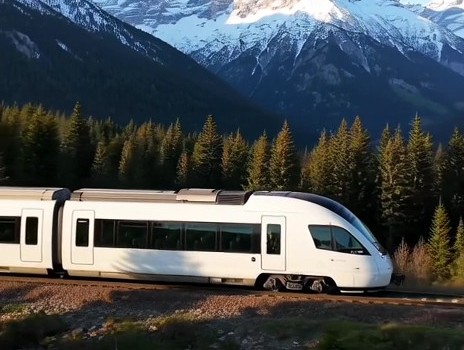}}%
  {\cleanpill}%
  {Train sounds\\[2pt]%
   {\fontsize{6.5}{8}\selectfont\textit{Clean -- corresponds to true anchor `train'}}}%
  {\cleanpill}%
  {\textit{``\ldots Clara unfolded the board, revealing a colorful map of North America
    crisscrossed with routes connecting major cities. Each player selected their colored
    pieces - tiny plastic carriages - and drew destination cards [\ldots]
    she had her eyes on the transcontinental path worth the most points\ldots''}\\[2pt]%
   {\fontsize{6.5}{8}\selectfont Passage describes Ticket to Ride board game -- clean, true anchor}}%
  {Train travels through a forested mountain landscape}%
\vspace{6pt}
{\fontsize{9}{11}\selectfont\textbf{Q:} The vehicle used for the journey present across the content travels which route?}
\vspace{4pt}
\optionblock{%
  \optOther{A}{Autobahn}\\[3pt]%
  \optPred{B}{Route 66}\enspace$\Leftarrow$\textbf{\textcolor{PredRed}{\fontsize{8}{9}\selectfont predicted}}\\[3pt]%
  \optAbs{C}{I cannot answer}\\[3pt]%
  \optOther{D}{Northwest Passage}\\[3pt]%
  \optGT{E}{Trans-Siberian Railway}\enspace\textcolor{GTGreen}{\qcmark~\textbf{\fontsize{8}{9}\selectfont human GT}}}
\vspace{5pt}
\begin{tcolorbox}[obox]
  {\fontsize{7}{8.5}\selectfont\textbf{Model output (MiniCPM, CoT):}}\vspace{3pt}\par
  {\fontsize{8}{10}\selectfont\textit{``The image shows a train crossing a bridge over a river.
    The train is black with a blue passenger car [\ldots] Route~66: This is a famous highway
    in the United States, but it is not a train route [\ldots] Trans-Siberian Railway: This
    is a long-distance railway line in Russia, not relevant to the North American context
    [\ldots] the most likely route is the transcontinental route [\ldots] The answer is~B.''
    \hfill$\Rightarrow$\,\textbf{predicts B}}}
\end{tcolorbox}
\vspace{4pt}
\begin{tcolorbox}[fbox2]
  {\fontsize{8.5}{10.5}\selectfont\textbf{\textcolor{PredRed}{Why it fails:}}\ All three
  modalities are clean and consistently reference trains. The model correctly identifies
  the anchor as a steam locomotive in the video and notes that the text describes a
  transcontinental route. It even \emph{explicitly rules out} Route~66, observing ``it is
  not a train route,'' yet selects it anyway. The question asks about real-world railway
  routes, not board-game geography; the Trans-Siberian Railway is the world's longest
  single rail line and the iconic exemplar of transcontinental train travel.
  \textbf{The model's own reasoning eliminates its chosen answer and correctly identifies
  the domain (transcontinental rail), yet it conflates the North American board-game
  setting with the real-world route the question targets - selecting a highway it has
  already dismissed over a railway it wrongly discards as geographically irrelevant.}}
\end{tcolorbox}
\qualboxend
\caption{E1: Wrong factual answer -- train ($k{=}0$, MiniCPM, synthetic).}
\label{fig:e1_train_clean}
\end{figure*}
\bigskip
\begin{figure*}[t]
\qualboxbegin{E1H}{E1Light}{E1\ \ Wrong Answer}%
  {$C_{001}$\enspace($k{=}1$, text corrupted)\enspace Anchor:\ \textit{car}}

\metabadge{Model: Gemini~2.5}\;
\metabadge{Prompt: zero-shot}\;
\metabadge{Split: synthetic}\;
\metabadge{True anchor: car}

\vspace{6pt}

\modalitygrid%
  {\cleanpill}%
  {\includegraphics[width=\linewidth]{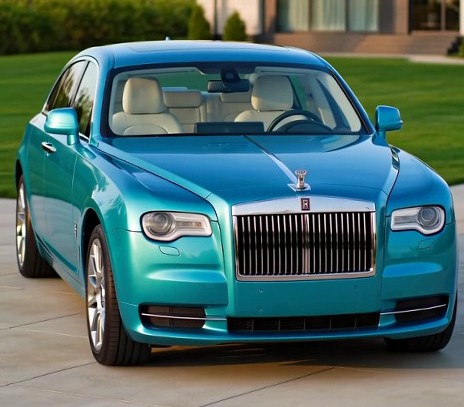}}%
  {\cleanpill}%
  {Car engine / driving sounds\\[2pt]%
   {\fontsize{6.5}{8}\selectfont\textit{Clean - corresponds to true anchor car'}}}%
  {\corruptpill}%
  {\textit{``\ldots massive bronze instruments hung silent [\ldots] the town would hear them
    ring again for Sunday service [\ldots] names of long-dead craftsmen\ldots''}\\[2pt]%
   {\fontsize{6.5}{8}\selectfont Passage describes church bells, not the true anchor car'}}%
  {Car driving on road}%

\vspace{6pt}
{\fontsize{9}{11}\selectfont\textbf{Q:} The object present across the content was first successfully developed by an engineer from which country?}

\vspace{4pt}
\optionblock{%
  \optPred{A}{Italy}\\[3pt]%
  \optOther{B}{England}\\[3pt]%
  \optGT{C}{Germany}\enspace\textcolor{GTGreen}{\qcmark~\textbf{\fontsize{8}{9}\selectfont human GT}}\\[3pt]%
  \optOther{D}{France}\\[3pt]%
  \optAbs{E}{I cannot answer}}

\vspace{5pt}
\begin{tcolorbox}[obox]
  {\fontsize{7}{8.5}\selectfont\textbf{Model output (Gemini~2.5, zero-shot):}}\vspace{3pt}\par
  {\fontsize{8}{10}\selectfont\textit{``The object described in the context is a bell,
    specifically a large bronze bell used for church services [\ldots]
    the earliest known bell foundries in Europe were in Italy, dating back to the 6th century
    AD. St.~Paulinus of Nola is often credited with introducing bells to Christian worship
    [\ldots]''
    \hfill$\Rightarrow$\,\textbf{predicts A}}}
\end{tcolorbox}

\vspace{4pt}
\begin{tcolorbox}[fbox2]
  {\fontsize{8.5}{10.5}\selectfont\textbf{\textcolor{PredRed}{Why it fails:}}\ Both clean
  modalities unambiguously show a car (video) and car engine sounds (audio), yet the model
  never mentions either. It reasons entirely from the corrupted bells text, identifies bells
  as the anchor, and answers a question about car engineering (Karl Benz, Germany) as if it
  were about medieval bell foundries in Italy.
  \textbf{The two clean modalities are completely ignored; a single corrupted text passage
  captures the model's anchor identification wholesale.}}
\end{tcolorbox}

\qualboxend
\caption{E1: Wrong factual answer - car ($k{=}1$, Gemini~2.5, synthetic).}
\label{fig:e1_car}
\end{figure*}


\begin{figure*}[t]
\subsubsection*{E2: Wrong Abstention}
\qualboxbegin{E2H}{E2Light}{E2\ \ False Abstention}%
  {$C_{101}$\enspace($k{=}2$, video+text corrupted)\enspace Anchor:\ \textit{horse}}

\metabadge{Model: Gemini~2.5}\;
\metabadge{Prompt: zero-shot}\;
\metabadge{Split: real}\;
\metabadge{True anchor: horse}

\vspace{6pt}

\modalitygrid%
  {\corruptpill}%
  {\includegraphics[width=\linewidth]{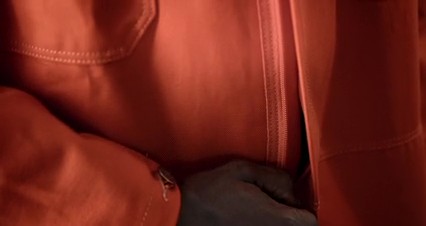}}%
  {\cleanpill}%
  {Horse neighing\\[2pt]%
   {\fontsize{6.5}{8}\selectfont\textit{Clean - corresponds to true anchor horse'}}}%
  {\corruptpill}%
  {\textit{``\ldots pinsetter working behind the lane [\ldots] triangular formation [\ldots] bowling\ldots''}\\[2pt]%
   {\fontsize{6.5}{8}\selectfont Passage describes bowling, not the true anchor horse'}}%
  {Zipper being pulled}%

\vspace{6pt}
{\fontsize{9}{11}\selectfont\textbf{Q:} The animal present across the content is the basis for which creature in Greek mythology?}

\vspace{4pt}
\optionblock{%
  \optAbs{A}{I cannot answer}\enspace\textcolor{PredRed}{\qxmark~\textbf{\fontsize{8}{9}\selectfont model predicts abstain}}\\[3pt]%
  \optGT{B}{Pegasus}\enspace\textcolor{GTGreen}{\qcmark~\textbf{\fontsize{8}{9}\selectfont human GT}}\\[3pt]%
  \optOther{C}{Cerberus}\\[3pt]%
  \optOther{D}{Chimera}\\[3pt]%
  \optOther{E}{Minotaur}}

\vspace{5pt}
\begin{tcolorbox}[obox]
  {\fontsize{7}{8.5}\selectfont\textbf{Model output (Gemini~2.5, zero-shot):}}\vspace{3pt}\par
  {\fontsize{8}{10}\selectfont\textit{``The audio in the video contains a distinct pig squeal [\ldots]
    Pegasus: A winged horse. The animal basis is a horse, not a pig [\ldots]
    Since the animal sound is a pig squeal, and none of the provided Greek mythological
    creatures are primarily based on a pig, the correct answer is that the question
    cannot be answered with the given options.''
    \hfill$\Rightarrow$\,\textbf{predicts A (abstain)}}}
\end{tcolorbox}

\vspace{4pt}
\begin{tcolorbox}[fbox2]
  {\fontsize{8.5}{10.5}\selectfont\textbf{\textcolor{PredRed}{Why it fails:}}\ The clean audio
  is an unambiguous horse neigh, yet the model explicitly claims to hear ``a distinct pig
  squeal'' - a hallucinated percept not attributable to either corrupted modality (zipper
  video; bowling text). The model then reasons correctly \emph{from its false percept}: since
  no listed Greek mythological creature is based on a pig, it abstains.
  \textbf{The error originates in spontaneous misidentification of a clean channel, not
  contamination from a corrupted one; the downstream reasoning is internally consistent
  but built on a fabricated foundation.}}
\end{tcolorbox}

\qualboxend
\caption{E2: Wrong abstention - horse ($k{=}2$, Gemini~2.5, real).}
\label{fig:e2_horse}
\end{figure*}

\bigskip

\begin{figure*}[t]
\qualboxbegin{E2H}{E2Light}{E2\ \ False Abstention}%
  {$C_{101}$\enspace($k{=}2$, video+text corrupted)\enspace Anchor:\ \textit{tiger}}

\metabadge{Model: Gemini~2.5}\;
\metabadge{Prompt: zero-shot}\;
\metabadge{Split: real}\;
\metabadge{True anchor: tiger}

\vspace{6pt}

\modalitygrid%
  {\corruptpill}%
  {\includegraphics[width=\linewidth]{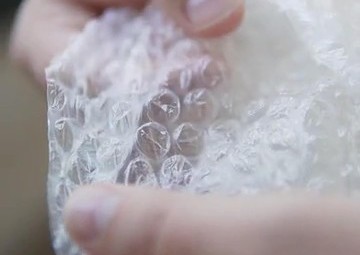}}%
  {\cleanpill}%
  {Tiger roar\\[2pt]%
   {\fontsize{6.5}{8}\selectfont\textit{Clean - corresponds to true anchor tiger'}}}%
  {\corruptpill}%
  {\textit{``\ldots the delivery room fell silent [\ldots] a sharp, wailing shriek as the
    tiny lungs filled with air for the first time\ldots''}\\[2pt]%
   {\fontsize{6.5}{8}\selectfont Passage describes a newborn crying, not the true anchor tiger'}}%
  {Bubble wrap being popped}%

\vspace{6pt}
{\fontsize{9}{11}\selectfont\textbf{Q:} The animal present across the content appears in the title of which film?}

\vspace{4pt}
\optionblock{%
  \optAbs{A}{I cannot answer}\enspace\textcolor{PredRed}{\qxmark~\textbf{\fontsize{8}{9}\selectfont model predicts abstain}}\\[3pt]%
  \optOther{B}{The Black Stallion}\\[3pt]%
  \optOther{C}{Enter the Dragon}\\[3pt]%
  \optOther{D}{Raging Bull}\\[3pt]%
  \optGT{E}{Crouching Tiger, Hidden Dragon}\enspace\textcolor{GTGreen}{\qcmark~\textbf{\fontsize{8}{9}\selectfont human GT}}}

\vspace{5pt}
\begin{tcolorbox}[obox]
  {\fontsize{7}{8.5}\selectfont\textbf{Model output (Gemini~2.5, zero-shot):}}\vspace{3pt}\par
  {\fontsize{8}{10}\selectfont\textit{``The video shows a finger popping bubble wrap. The audio
    contains the distinct sound of a baby crying [\ldots] None of the film titles (Stallion,
    Dragon, Bull, Tiger) correspond to a human baby or any other animal that could be
    reasonably inferred from the video or audio [\ldots] the question cannot be answered.''
    \hfill$\Rightarrow$\,\textbf{predicts A (abstain)}}}
\end{tcolorbox}

\vspace{4pt}
\begin{tcolorbox}[fbox2]
  {\fontsize{8.5}{10.5}\selectfont\textbf{\textcolor{PredRed}{Why it fails:}}\ The clean audio
  is an unambiguous tiger roar, yet the model explicitly describes it as ``the distinct sound
  of a baby crying'' - a misperception directly induced by the corrupted text, which describes
  a newborn's first cry. The corrupted passage primes the model's interpretation of the clean
  audio channel, causing it to hear a baby instead of a tiger.
  \textbf{Once the audio is misread, the downstream reasoning is internally consistent: no
  film title matches a baby, so the model correctly abstains - from a fabricated premise.}}
\end{tcolorbox}

\qualboxend
\caption{E2: Wrong abstention - tiger ($k{=}2$, Gemini~2.5, synthetic).}
\label{fig:e2_tiger}
\end{figure*}

\bigskip

\begin{figure*}[t]
\qualboxbegin{E2H}{E2Light}{E2\ \ False Abstention}%
  {$C_{000}$\enspace($k{=}0$, all clean)\enspace Anchor:\ \textit{typewriter}}

\metabadge{Model: GPT-4o-Mini}\;
\metabadge{Prompt: CoT}\;
\metabadge{Split: synthetic}\;
\metabadge{True anchor: typewriter}

\vspace{6pt}

\modalitygrid%
  {\cleanpill}%
  {\includegraphics[width=\linewidth]{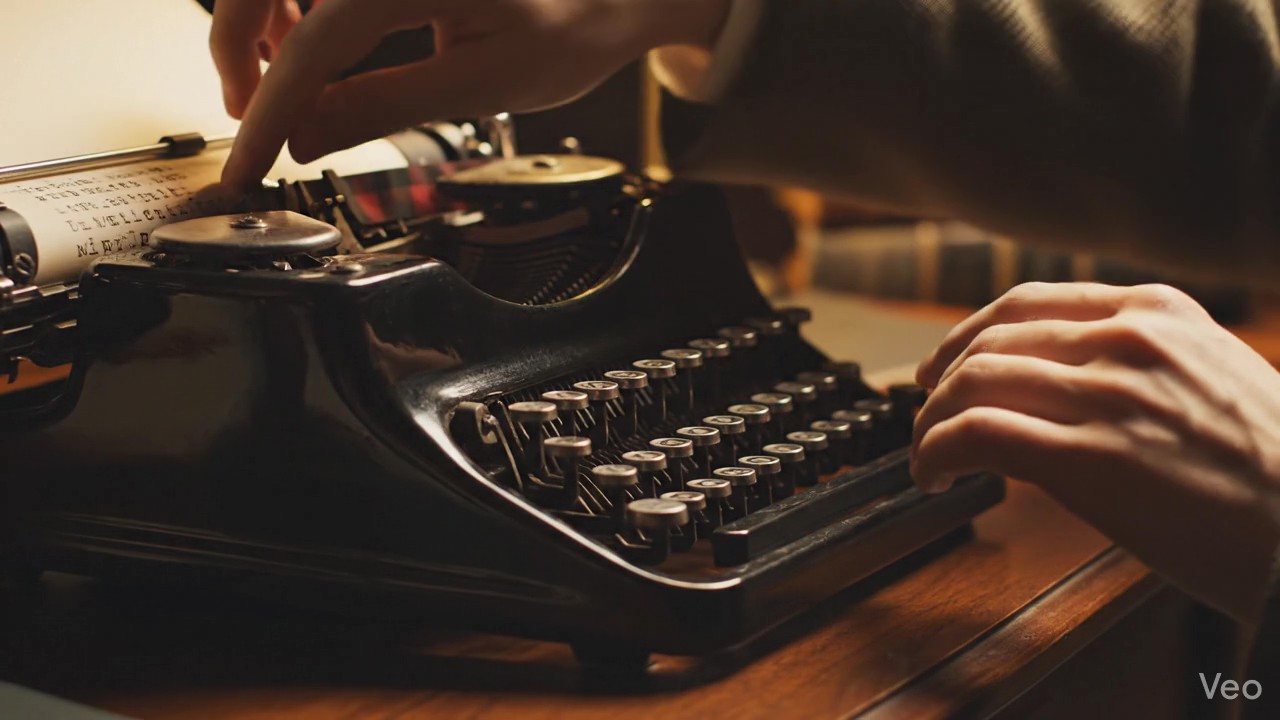}}%
  {\cleanpill}%
  {Typewriter keys clacking\\[2pt]%
   {\fontsize{6.5}{8}\selectfont\textit{True anchor - matches `typewriter'}}}%
  {\cleanpill}%
  {\textit{``\ldots pressing a key sent a corresponding hammer upward to strike the
    ribbon against the platen [\ldots] the same machine Hemingway had used,
    the shopkeeper claimed\ldots''}\\[2pt]%
   {\fontsize{6.5}{8}\selectfont Passage describes a typewriter - clean, true anchor}}%
  {Typewriter being used}%

\vspace{6pt}
{\fontsize{9}{11}\selectfont\textbf{Q:} The device present across the content features prominently in which film?}

\vspace{4pt}
\optionblock{%
  \optPred{A}{The Shining}\enspace\ldots\enspace\textcolor{GTGreen}{\qcmark~\textbf{\fontsize{8}{9}\selectfont human GT}}\\[3pt]%
  \optAbs{B}{I cannot answer}\enspace$\Leftarrow$\textbf{\textcolor{PredRed}{\fontsize{8}{9}\selectfont predicted}}\\[3pt]%
  \optOther{C}{The Machinist}\\[3pt]%
  \optOther{D}{The Imitation Game}\\[3pt]%
  \optOther{E}{The Pianist}}

\vspace{5pt}
\begin{tcolorbox}[obox]
  {\fontsize{7}{8.5}\selectfont\textbf{Model output (GPT-4o-Mini, CoT):}}\vspace{3pt}\par
  {\fontsize{8}{10}\selectfont\textit{``The Shining: This film does not prominently feature a typewriter [\ldots]
    The Machinist: while it has a dark theme, it does not focus on a typewriter [\ldots]
    The Imitation Game: centers around code-breaking and features machines, but not
    specifically a typewriter as a central element [\ldots]
    none of the options directly relate to a typewriter in a significant way.''
    \hfill$\Rightarrow$\,\textbf{predicts B}}}
\end{tcolorbox}

\vspace{4pt}
\begin{tcolorbox}[fbox2]
  {\fontsize{8.5}{10.5}\selectfont\textbf{\textcolor{PredRed}{Why it fails:}}\ All three
  modalities are clean and consistently identify a typewriter. Jack Torrance's typewriter
  in \textit{The Shining} is among cinema's most iconic props - the source of the
  endlessly repeated ``All work and no play'' manuscript. The model correctly identifies
  the anchor from the multimodal content but then explicitly and incorrectly states that
  \textit{The Shining} does not prominently feature a typewriter, abstaining on the basis
  of a factual knowledge error alone. \textbf{This is a zero-corruption knowledge failure:
  no corrupted channel, no cross-modal confusion - the model simply holds a wrong belief
  about a well-known film and treats that belief as sufficient grounds to abstain.}}
\end{tcolorbox}

\qualboxend
\caption{E2: False abstention - typewriter ($k{=}0$, GPT-4o-Mini, synthetic).}
\label{fig:e2_typewriter}
\end{figure*}

\bigskip


\begin{figure*}[t]
\subsubsection*{E3: Missed Abstention}
\qualboxbegin{E3H}{E3Light}{E3\ \ Missed Abstention}%
  {$C_{111}$\enspace($k{=}3$, all corrupted)\enspace Anchor:\ \textit{piano}}

\metabadge{Model: Qwen3o-Instruct}\;
\metabadge{Prompt: CoT}\;
\metabadge{Split: real}\;
\metabadge{True anchor: piano}

\vspace{6pt}

\modalitygrid%
  {\corruptpill}%
  {\includegraphics[width=\linewidth]{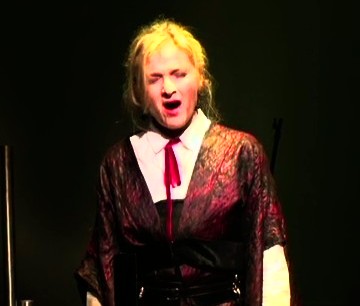}}%
  {\corruptpill}%
  {Kettle whistle / steam hiss\\[2pt]%
   {\fontsize{6.5}{8}\selectfont\textit{From a different anchor - not piano'}}}%
  {\corruptpill}%
  {\textit{``\ldots pedestrians hurried past with umbrellas overhead [\ldots]
    taxi drivers cursed the reduced visibility [\ldots] storm drains gurgled\ldots''}\\[2pt]%
   {\fontsize{6.5}{8}\selectfont Passage describes rain, not the true anchor piano'}}%
  {Opera performance}%

\vspace{6pt}
{\fontsize{9}{11}\selectfont\textbf{Q:} The instrument present across the content is most associated with which 19th-century virtuoso?}

\vspace{4pt}
\optionblock{%
  \optOther{A}{Pablo Casals}\\[3pt]%
  \optOther{B}{Franz Liszt}\\[3pt]%
  \optAbs{C}{I cannot answer}\enspace\textcolor{GTGreen}{\qcmark~\textbf{\fontsize{8}{9}\selectfont human GT}}\\[3pt]%
  \optOther{D}{Andr\'{e}s Segovia}\\[3pt]%
  \optPred{E}{Niccol\`{o} Paganini}}

\vspace{5pt}
\begin{tcolorbox}[obox]
  {\fontsize{7}{8.5}\selectfont\textbf{Model output (Qwen3o-Instruct, CoT):}}\vspace{3pt}\par
  {\fontsize{8}{10}\selectfont\textit{``The audio clip features a very high-pitched, sustained,
    and vibrato-rich sound, characteristic of a bowed string instrument [\ldots]
    extremely high in pitch, suggesting a smaller instrument like a piccolo violin
    or a viola d'amore [\ldots] most consistent with the kind of virtuosic,
    high-register playing that Paganini was famous for.''
    \hfill$\Rightarrow$\,\textbf{predicts E}}}
\end{tcolorbox}

\vspace{4pt}
\begin{tcolorbox}[fbox2]
  {\fontsize{8.5}{10.5}\selectfont\textbf{\textcolor{PredRed}{Why it fails:}}\ All three
  modalities are corrupted (opera video; kettle whistle audio; rain text), yet the model
  builds a confident musicological chain entirely from the corrupted audio: it describes the
  kettle whistle as a \textit{``vibrato-rich bowed string instrument''} and proceeds to
  identify Paganini's harmonic techniques as the closest match.
  \textbf{No cross-modal consistency check is performed; a tea kettle is confabulated
  into a virtuoso violin performance and committed to as a factual answer.}}
\end{tcolorbox}

\qualboxend
\caption{E3: Missed abstention - piano ($k{=}3$, Qwen3o-Instruct, real).}
\label{fig:e3_piano_qwen}
\end{figure*}

\bigskip

\begin{figure*}[t]
\qualboxbegin{E3H}{E3Light}{E3\ \ Missed Abstention}%
  {$C_{110}$\enspace($k{=}2$, video+audio corrupted)\enspace Anchor:\ \textit{stapler}}

\metabadge{Model: GPT-4o-Mini}\;
\metabadge{Prompt: CoT}\;
\metabadge{Split: real}\;
\metabadge{True anchor: stapler}

\vspace{6pt}

\modalitygrid%
  {\corruptpill}%
  {\includegraphics[width=\linewidth]{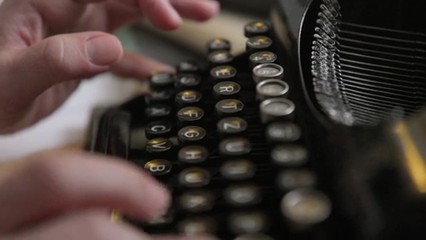}}%
  {\corruptpill}%
  {Train horn blast\\[2pt]%
   {\fontsize{6.5}{8}\selectfont\textit{From a different anchor - not `stapler'}}}%
  {\cleanpill}%
  {\textit{``\ldots grabbed the small metal device beside her keyboard [\ldots]
    pressed down firmly. A sharp click echoed as the thin metal teeth
    punctured through the sheets\ldots''}\\[2pt]%
   {\fontsize{6.5}{8}\selectfont Passage describes stapler - clean, true anchor}}%
  {Typewriter being used}%

\vspace{6pt}
{\fontsize{9}{11}\selectfont\textbf{Q:} The tool present across the content became a pop culture icon through which film?}

\vspace{4pt}
\optionblock{%
  \optAbs{A}{I cannot answer}\enspace\textcolor{GTGreen}{\qcmark~\textbf{\fontsize{8}{9}\selectfont human GT}}\\[3pt]%
  \optPred{B}{Office Space}\\[3pt]%
  \optOther{C}{The Shining}\\[3pt]%
  \optOther{D}{American Psycho}\\[3pt]%
  \optOther{E}{Secretary}}

\vspace{5pt}
\begin{tcolorbox}[obox]
  {\fontsize{7}{8.5}\selectfont\textbf{Model output (GPT-4o-Mini, CoT):}}\vspace{3pt}\par
  {\fontsize{8}{10}\selectfont\textit{``Linda is using a metal device to bind loose pages together,
    which suggests a stapler [\ldots]
    Office Space: This film features a stapler prominently, especially the character
    Milton's attachment to his red stapler [\ldots]
    the most fitting answer is Office Space.''
    \hfill$\Rightarrow$\,\textbf{predicts B}}}
\end{tcolorbox}

\vspace{4pt}
\begin{tcolorbox}[fbox2]
  {\fontsize{8.5}{10.5}\selectfont\textbf{\textcolor{PredRed}{Why it fails:}}\ The model
  correctly identifies the anchor from the single clean text modality and retrieves the
  right cultural association. However, with both video (typewriter) and audio (train horn)
  corrupted and showing entirely unrelated content, a consistent cross-modal identity signal
  is absent - the model should have detected this inconsistency and abstained. Typerwriter also strongly supports The Shining (Option C).
  \textbf{Anchoring entirely on one clean channel while ignoring two corrupted contradictors
  produces a confident correct-seeming answer when the benchmark requires abstention.}}
\end{tcolorbox}

\qualboxend
\caption{E3: Missed abstention - stapler ($k{=}2$, GPT-4o-Mini, real).}
\label{fig:e3_stapler}
\end{figure*}

\bigskip

\begin{figure*}[t]
\qualboxbegin{E3H}{E3Light}{E3\ \ Missed Abstention}%
  {$C_{111}$\enspace($k{=}3$, all corrupted)\enspace Anchor:\ \textit{typewriter}}

\metabadge{Model: Qwen3o-Instruct}\;
\metabadge{Prompt: CoT}\;
\metabadge{Split: real}\;
\metabadge{True anchor: typewriter}

\vspace{6pt}

\modalitygrid%
  {\corruptpill}%
  {\includegraphics[width=\linewidth]{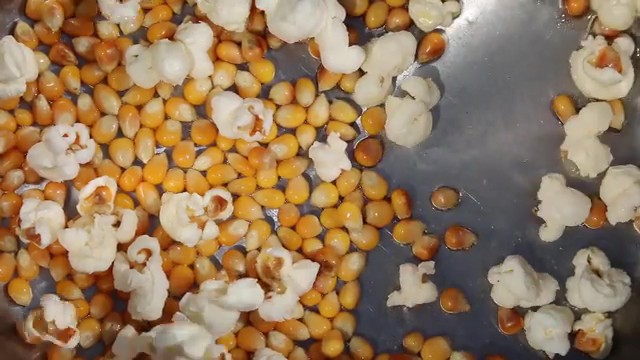}}%
  {\corruptpill}%
  {Clock ticking\\[2pt]%
   {\fontsize{6.5}{8}\selectfont\textit{From a different anchor - not typewriter'}}}%
  {\corruptpill}%
  {\textit{``\ldots calculate the maximum expected intensity [\ldots]
    heaviest precipitation event likely to occur over a fifty-year period [\ldots]
    water accumulation rates measured in millimetres per hour\ldots''}\\[2pt]%
   {\fontsize{6.5}{8}\selectfont Passage describes roof drainage engineering, not typewriter'}}%
  {Popcorn kernels popping}%

\vspace{6pt}
{\fontsize{9}{11}\selectfont\textbf{Q:} The device present across the content features prominently in which film?}

\vspace{4pt}
\optionblock{%
  \optOther{A}{The Machinist}\\[3pt]%
  \optOther{B}{The Imitation Game}\\[3pt]%
  \optAbs{C}{I cannot answer}\enspace\textcolor{GTGreen}{\qcmark~\textbf{\fontsize{8}{9}\selectfont human GT}}\\[3pt]%
  \optPred{D}{The Shining}\\[3pt]%
  \optOther{E}{The Pianist}}

\vspace{5pt}
\begin{tcolorbox}[obox]
  {\fontsize{7}{8.5}\selectfont\textbf{Model output (Qwen3o-Instruct, CoT):}}\vspace{3pt}\par
  {\fontsize{8}{10}\selectfont\textit{`The video shows a close-up of popcorn kernels being
    popped in a pan. This is a classic scene from Stanley Kubrick's film The Shining.'
    In The Shining,' the character Jack Torrance is shown popping popcorn in a pan while
    he is in the Overlook Hotel's kitchen. The sound of the kernels popping is a key
    auditory element in that scene [\ldots] the device is the pan used for popping popcorn,
    which is a central element in the scene from The Shining.'\,''
    \hfill$\Rightarrow$\,\textbf{predicts D}}}
\end{tcolorbox}

\vspace{4pt}
\begin{tcolorbox}[fbox2]
  {\fontsize{8.5}{10.5}\selectfont\textbf{\textcolor{PredRed}{Why it fails:}}\ The model
  invents a popcorn-popping scene in \textit{The Shining} that does not exist in that
  film - a pure hallucination triggered by the corrupted video. It then constructs a
  self-consistent but entirely fabricated narrative (Jack Torrance, Overlook Hotel kitchen,
  popping sounds as ``key auditory element'') to justify its answer.
  \textbf{With all three modalities corrupted and no coherent cross-modal signal available,
  the model confabulates a plausible film-scene memory rather than recognising the
  absence of a consistent anchor and abstaining.}}
\end{tcolorbox}

\qualboxend
\caption{E3: Missed abstention - typewriter ($k{=}3$, Qwen3o-Instruct, real).}
\label{fig:e3_typewriter_qwen}
\end{figure*}

\bigskip

\begin{figure*}[t]
\qualboxbegin{E3H}{E3Light}{E3\ \ Missed Abstention}%
  {$C_{111}$\enspace($k{=}3$, all corrupted)\enspace Anchor:\ \textit{cat}}

\metabadge{Model: MiniCPM}\;
\metabadge{Prompt: CoT}\;
\metabadge{Split: synthetic}\;
\metabadge{True anchor: cat}

\vspace{6pt}

\modalitygrid%
  {\corruptpill}%
  {\includegraphics[width=\linewidth]{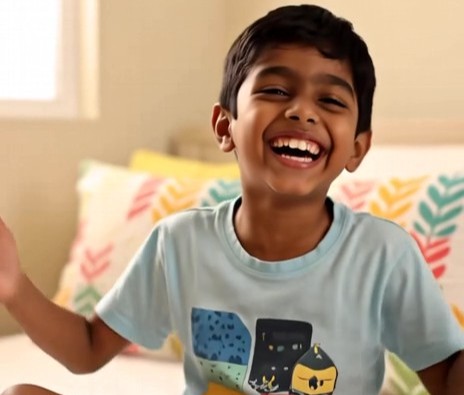}}
  {\corruptpill}%
  {Tiger roar\\[2pt]%
   {\fontsize{6.5}{8}\selectfont\textit{From a different anchor - not cat'}}}%
  {\corruptpill}%
  {\textit{``\ldots the beautiful dark-coated creature narrated its journey through
    Victorian England, pulling carriages [\ldots] the book had sparked early animal
    welfare movements when it was published in 1877\ldots''}\\[2pt]%
   {\fontsize{6.5}{8}\selectfont Passage is Black Beauty - describes a horse, not cat'}}%
  {Child laughing}%

\vspace{6pt}
{\fontsize{9}{11}\selectfont\textbf{Q:} The creature present across the content lends its name to which term for a short daytime sleep?}

\vspace{4pt}
\optionblock{%
  \optOther{A}{Dog tired}\\[3pt]%
  \optOther{B}{Bird rest}\\[3pt]%
  \optAbs{C}{I cannot answer}\enspace\textcolor{GTGreen}{\qcmark~\textbf{\fontsize{8}{9}\selectfont human GT}}\\[3pt]%
  \optPred{D}{Cat nap}\\[3pt]%
  \optOther{E}{Bear hibernation}}

\vspace{5pt}
\begin{tcolorbox}[obox]
  {\fontsize{7}{8.5}\selectfont\textbf{Model output (MiniCPM, CoT):}}\vspace{3pt}\par
  {\fontsize{8}{10}\selectfont\textit{`The story is about an animal that was born on an English farm,
    sold to various owners, and experienced different conditions [\ldots]
    the animal in question is likely a cat, as the context mentions a creature
    that was born on a farm and experienced various conditions, which aligns
    with the life of a cat [\ldots] the term that best fits the description of
    a short daytime sleep associated with a cat is cat nap.'\,''
    \hfill$\Rightarrow$\,\textbf{predicts D}}}
\end{tcolorbox}

\vspace{4pt}
\begin{tcolorbox}[fbox2]
  {\fontsize{8.5}{10.5}\selectfont\textbf{\textcolor{PredRed}{Why it fails:}}\ The corrupted
  text is the opening of \textit{Black Beauty} - an unambiguous description of a horse
  (Victorian England, carriage work, 1877 publication). All three modalities are corrupted
  and none contains a cat. Yet the model reads the Black Beauty passage and concludes the
  anchor is a cat on the basis that ``born on a farm and experienced various conditions
  aligns with the life of a cat.'' \textbf{Rather than detecting the absence of a
  consistent cross-modal anchor, the model confabulates a plausible-sounding
  but factually wrong species identification from the corrupted text alone, then
  confidently answers from that hallucinated premise.}}
\end{tcolorbox}

\qualboxend
\caption{E3: Missed abstention - cat ($k{=}3$, MiniCPM, synthetic).}
\label{fig:e3_cat_minicpm}
\end{figure*}

\end{document}